%% file: main.tex
\documentclass{article}

\usepackage{graphicx}
\graphicspath{{figures/}}

\usepackage{arxiv}
\usepackage[utf8]{inputenc}
\usepackage[T1]{fontenc}

\usepackage{amsmath}
\usepackage{amssymb}
\usepackage{amsfonts}
\usepackage{nicefrac}

\usepackage{enumitem}
\usepackage{booktabs}
\usepackage{tabularx}
\usepackage{multirow}
\usepackage{array}
\usepackage{float}
\usepackage{fancyvrb}
\usepackage{placeins}

\usepackage{caption}
\usepackage{subcaption}
\renewcommand{\arraystretch}{1.15}

\usepackage{xcolor}
\usepackage{tikz}
\usetikzlibrary{positioning,arrows.meta,shapes.geometric,calc,fit,backgrounds}

\usepackage[colorlinks, linkcolor=blue, citecolor=blue, urlcolor=blue]{hyperref}
\usepackage{url}

\usepackage[numbers,sort&compress]{natbib}
\usepackage{microtype}

\newcolumntype{L}[1]{>{\raggedright\let\newline\\\arraybackslash\hspace{0pt}}m{#1}}
\newcolumntype{C}[1]{>{\centering}m{#1}}
\newcolumntype{R}[1]{>{\raggedleft\let\newline\\\arraybackslash\hspace{0pt}}m{#1}}

\definecolor{ao}{rgb}{0.0, 0.0, 1.0}

\title{\textbf{AniMatrix}: An Anime Video Generation Model that Thinks in Art, Not Physics}

\author{Tencent HY Team}

\begin{document}
\maketitle

% -----------------------------------------------
%  摘要
% -----------------------------------------------
\begin{abstract}
\input{abstract}
\end{abstract}

% ================================================================
\section{Introduction}
\label{sec:intro}
\input{intro_v2}

% ================================================================
\section{Related Work}
\label{sec:related}
\input{related_work}

% ================================================================
\section{Data Preparation}
\label{sec:data}
\input{data_preparation}

% ================================================================
\section{Model Design}
\label{sec:model}
\input{model_design}

% ================================================================
\section{Training Strategy}
\label{sec:training}
\input{training_strategy}

% ================================================================
\section{Experiments}
\label{sec:eval}
\input{evaluation}

% ================================================================
\section{Inference and Deployment}
\label{sec:deploy}
\input{inference_deployment}

% ================================================================
\section{Conclusion}
\label{sec:conclusion}
\input{conclusions}

% ================================================================
\section{Contributors}
\label{sec:contributors}
\input{contributors}

% ================================================================
%  参考文献
% ================================================================
\bibliography{references}

% ================================================================
%  附录（Supplementary Material）
% ================================================================
% Force the appendix to start on a fresh page so the bibliography list does
% not visually flow into "A Supplementary Material...".
\clearpage
\appendix
% In the appendix, force \subsection (A.1/A.2/A.3) to flush pending floats via
% \FloatBarrier. This blocks tables from drifting across the three top-level
% appendix blocks (Taxonomy, AniCaption, Data Curation), so e.g. Tables 10--11
% (A.1) cannot land after Table 12 (A.2). \subsubsection is left unbarriered so
% that figures and tables can flow naturally inside each block and fill page
% bottoms instead of leaving 60\%+ blank pages at A.2.x boundaries.
\let\AniMatrixOldSubsection\subsection
\renewcommand{\subsection}{\FloatBarrier\AniMatrixOldSubsection}

\section{Supplementary Material for Data Preparation}
\label{app:data_supp}

The supplementary material expands the detailed evidence behind the
compressed Data Preparation discussion in Section~\ref{sec:data}. The
three subsections correspond one-to-one to the three pillars of
Section~\ref{sec:data}:
Appendix~\ref{app:taxonomy} expands the Industrial Production
Taxonomy (Sec.~\ref{sec:data:taxonomy}); Appendix~\ref{app:anicaption}
covers the AniCaption schema, training, and evaluation
(Sec.~\ref{sec:data:caption}); Appendix~\ref{app:data_curation_section}
gives the data curation operators and expert rubric
(Sec.~\ref{sec:data:filtering}).

\subsection{Industrial Production Taxonomy: Detailed Vocabularies
            \texorpdfstring{(Supplements Sec.~\ref{sec:data:taxonomy})}{}}
\label{app:taxonomy}
\input{appendix_taxonomy}

\subsection{AniCaption: Schema, Training, and Evaluation Protocols
            \texorpdfstring{(Supplements Sec.~\ref{sec:data:caption})}{}}
\label{app:anicaption}
\input{appendix_anicaption}

\subsection{Data Curation: Operators, Anime-Specific Filters, and Expert Rubric
            \texorpdfstring{(Supplements Sec.~\ref{sec:data:filtering})}{}}
\label{app:data_curation_section}
\input{appendix_data_curation}

\end{document}

%% file: abstract.tex
Video generation models internalize physical realism as their prior. Anime deliberately violates physics---smears, impact frames, chibi shifts---and its thousands of coexisting artistic conventions yield no single ``physics of anime'' a model can absorb. Physics-biased models therefore flatten the artistry that defines the medium or collapse under its stylistic variance.
We present AniMatrix, a video generation model that makes artistic correctness the optimization target through a dual-channel conditioning interface and a three-step transition: \emph{redefine correctness, override the physics prior, and distinguish art from failure.}
First, a Production Knowledge System encodes anime as a structured taxonomy of controllable production variables---Style $\times$ Motion $\times$ Camera $\times$ VFX---and AniCaption infers these variables from pixels as directorial directives. A dual-channel conditioning interface separates structured production tags from free-form narrative, keeping categorical directives enforceable during generation.
Second, a style--motion--deformation curriculum transitions the model from near-physical motion to expressive anime motion.
Third, deformation-aware preference optimization with a domain-specific reward model separates intentional artistry from pathological collapse.
 On an anime-specific human evaluation with five production dimensions scored by professional animators, AniMatrix ranks first on four of five, with the largest gains over Seedance-Pro~1.0~\cite{seedance2024} on Prompt Understanding ($+$0.70, $+$22.4\%) and Artistic Motion ($+$0.55, $+$16.9\%). We are preparing accompanying resources for public release to support reproducibility and follow-up research.

%% file: intro_v2.tex
Video generation has advanced rapidly, with models such as Sora~\cite{brooks2024sora}, HunyuanVideo~\cite{kong2024hunyuanvideo}, Wan 2.2~\cite{wan2025}, Kling~\cite{klingomni2025}, CogVideoX~\cite{yang2024cogvideox}, Seedance~\cite{seedance2025pro}, and SkyReels~\cite{skyreelsv2,skyreelsv4} producing coherent and visually rich natural video.
A key reason for this progress is that natural video obeys a single, universal set of physical laws: gravity pulls objects downward, rigid bodies conserve momentum, and light scatters according to well-defined optics.
Every frame of natural video data implicitly encodes these laws, providing a structural prior that diffusion models~\cite{ho2020ddpm,rombach2022ldm,peebles2023dit,blattmann2023videoldm,lipman2023flow} learn implicitly during training.

Anime operates under a fundamentally different paradigm.
Its motion vocabulary is not a noisy approximation of physics but a deliberate departure from it.
The classical principles of animation---squash-and-stretch, anticipation, exaggeration, timing~\cite{thomas1981illusion}---are precisely the places where anime \emph{intentionally violates} physical law for dramatic, emotional, and rhythmic effect.
A character leaping in anime does not follow a parabolic arc.
It squeezes in anticipation, stretches explosively, hangs mid-air for dramatic emphasis, and lands with an impact frame that shatters perspective.
But motion exaggeration is only one dimension of anime's departure from physics.
A series may shift abruptly from intense combat to chibi proportions---not because the drawing ``broke,'' but because the director uses tonal contrast to release dramatic tension.
These techniques---tonal shifts, exaggeration, rhythmic holds---have no counterpart in physical reality, yet each reflects a precise directorial intention.
They are not noise to be filtered---they are the art form.

This distinction exposes a deeper structural problem.
The physical world, despite its visual complexity, is governed by a compact set of universal laws that impose strong regularity on all data.
A model trained on any subset of natural video will converge toward the same implicit physical prior, because every sample obeys the same rules.
Anime has no such universal law.
Its artistic space is vast and stylistically diverse (from Miyazaki's painterly naturalism to Imaishi's kinetic maximalism), and inherently impressionistic---the same emotion can be conveyed through entirely different visual strategies across studios, directors, and eras.
Thousands of coexisting artistic conventions, each with its own internal logic, produce a training signal that is too diverse and mutually contradictory for any single implicit prior to emerge automatically.

Current attempts to generate anime video inherit the physics-first assumption in one of three ways.
The most direct route---fine-tuning general-purpose video models on anime data---retains the physics prior and produces motion that is technically smooth but artistically flat, suppressing the exaggeration that defines anime.
An alternative is to scale anime data na\"ively, but the model cannot reconcile the extreme variance of artistic styles with its physics-trained internal representations, leading to early collapse.
A third line trains with descriptive captions that record what is visible rather than what production decisions to follow, improving semantic coverage but still treating anime as an observed result to reconstruct.

AniSora~\cite{jiang2024anisora} explores controllable anime generation with a spatiotemporal mask module, and AnimeReward~\cite{zhu2025animereward} proposes human-feedback alignment for anime video quality.
Both adapt physics-trained models to anime stylistically without reconsidering whether physical realism remains the correct optimization target.
For anime, it is not.

We present AniMatrix, a video generation model that targets artistic rather than physical correctness.
AniMatrix makes this objective architecturally concrete through a dual-channel conditioning mechanism that separates precise production control from flexible artistic intent.
It makes the objective operationally concrete through a three-step transition that answers one question: \emph{how does a model transition from physical correctness to artistic correctness?}

The first step is to \textbf{redefine what ``correct'' means}.
We construct a \emph{Production Knowledge System} to supply the organizing structure that anime data alone cannot provide.
It comprises an Industrial Production Taxonomy that factorizes anime clips into four controllable production-variable axes---\textbf{Style}, \textbf{Motion}, \textbf{Camera}, and \textbf{VFX}---together with a dedicated annotation system (AniCaption) augmented by graph-based multimodal reasoning~\cite{qwen3vl}, and a dual-channel creator-language conditioning scheme.
AniCaption is not merely a captioner: it infers these production variables from pixels and verbalizes them as directorial directives.
Conditioning on them therefore specifies how a clip should be authored, not merely what should appear in it.
Correctness shifts from reconstructing visible content under a physically plausible prior to realizing a specified production plan.

Once we define this new objective, the second step is to \textbf{progressively override the physics prior}.
The distributional distance between a physics prior and extreme anime expression is too large for direct training, as high-deformation content combined with diverse style distributions causes early collapse~\cite{bengio2009curriculum,kim2024curriculum_diffusion}.
We address this with a progressive \emph{style--motion--deformation curriculum} whose three axes---style diversity, motion amplitude, and deformation intensity---each correspond to a distinct way anime departs from physics.
The curriculum provides a controlled transition from near-physical motion to extreme artistic expression and prevents the collapse that plagues na\"ive mixed training.

Finally, once the model produces exaggerated motion, the third step is to \textbf{distinguish art from failure}.
Both intentional artistic expression and pathological structural collapse appear as ``physics violations'' to standard metrics such as FVD and CLIP score.
We introduce \emph{deformation-aware preference optimization}~\cite{rafailov2023dpo,videodpo2024} with a domain-specific reward model that establishes a new quality standard within the artistic-correctness paradigm, separating intentional artistry from pathological collapse.

These three stages form a closed loop---define the new objective, learn to reach it, and distinguish right from wrong under it---and together constitute our main contributions:
\begin{itemize}[nosep,leftmargin=*]
\item \textbf{Redefining ``correctness.''} A Production Knowledge System encodes anime production knowledge as a structured taxonomy ($\mathcal{S}\!\times\!\mathcal{M}\!\times\!\mathcal{C}\!\times\!\mathcal{V}$), and AniCaption~\cite{qwen3vl} infers these variables from pixels as directorial directives.
\item \textbf{Architecture, curriculum, and alignment for artistic correctness.}
A trainable tag encoder preserves the orthogonal field--value structure of the taxonomy, while a frozen umT5-XXL encoder handles free-form narrative.
Dual-path injection feeds both channels into the generator---cross-attention for fine-grained per-token control and AdaLN modulation for global production-attribute enforcement at every layer---so that categorical directives are never diluted by open-ended text.
A style--motion--deformation curriculum then provides a controlled transition from near-physical motion to full anime expressiveness, with each axis corresponding to a distinct way anime departs from physics.
A domain-specific reward model drives deformation-aware preference optimization~\cite{rafailov2023dpo,videodpo2024}, establishing a quality boundary that separates intentional artistry from pathological collapse.
\item \textbf{Anime-specific evaluation and results.}
A five-dimension human evaluation framework scored by professional animators replaces physics-calibrated metrics that anti-correlate with anime quality.
AniMatrix ranks first on four of five dimensions, with the largest gains over Seedance-Pro~1.0~\cite{seedance2024} on Prompt Understanding ($+$22.4\%) and Artistic Motion ($+$16.9\%).
\end{itemize}

%% file: related_work.tex
\subsection{Video Generative Models}

Diffusion models~\cite{ho2020ddpm,song2021scorebased,song2021ddim,dhariwal2021diffusionbeatsgan} have transformed video generation, evolving from early U-Net-based architectures~\cite{rombach2022ldm,blattmann2023videoldm} to scalable Diffusion Transformer (DiT) frameworks~\cite{peebles2023dit,vaswani2017attention} operating in compressed latent spaces, with Flow Matching~\cite{lipman2023flow} further improving convergence and sample quality.

Sora~\cite{brooks2024sora} demonstrated the effectiveness of large-scale DiT training with spatiotemporal attention for generating coherent, temporally extended videos. In the open-source domain, models such as HunyuanVideo~\cite{kong2024hunyuanvideo}, Wan 2.2~\cite{wan2025}, CogVideoX~\cite{yang2024cogvideox}, Open-Sora~\cite{opensora2024}, and SkyReels~\cite{skyreelsv2,skyreelsv4} have rapidly narrowed the gap with proprietary systems such as Kling~\cite{klingomni2025}, Seedance~\cite{seedance2024,seedance2025pro}, and Vidu~\cite{vidu2024} through scaling data curation~\cite{chen2024panda70m} and architectural improvements including 3D RoPE~\cite{su2024rope}, Mixture-of-Experts~\cite{fedus2022switch}, and efficient attention~\cite{dao2023flashattention2}, with standard benchmarks such as FVD~\cite{unterthiner2019fvd}, FID~\cite{heusel2017fid}, and VBench~\cite{huang2024vbench} tracking this progress. The effectiveness of this paradigm rests on a single, often unstated premise: natural video implicitly encodes a universal physical prior that diffusion models absorb automatically during training. Despite this progress, all of these models are trained on natural video corpora and optimize for physical realism---an objective fundamentally misaligned with professional anime generation, where \emph{artistic correctness}, not physical fidelity, is the target.

\subsection{Anime Video Generation and Its Unique Challenges}

AniSora~\cite{jiang2024anisora} introduces a spatiotemporal mask module for controllable anime generation, while AnimeReward~\cite{zhu2025animereward} proposes human-feedback alignment specifically calibrated for anime video quality. ToonCrafter~\cite{xing2024tooncrafter} addresses the related problem of interpolating between anime keyframes, and AnimateDiff~\cite{guo2024animatediff} enables motion customization for stylized animation through lightweight temporal adapters. A common limitation is that they adapt physics-trained models to anime stylistically without reconsidering whether physical realism itself remains the correct optimization target. AniMatrix addresses all three stages---redefining correctness, unlearning the physics prior, and distinguishing art from failure---rather than adapting a single component of an otherwise physics-oriented pipeline.

\subsection{Preference Optimization for Generative Models}

Direct Preference Optimization (DPO)~\cite{rafailov2023dpo} provides an offline alternative to reinforcement learning from human feedback (RLHF)~\cite{ouyang2022instructgpt}, optimizing directly on preference pairs without requiring a separate reward model during training. VideoDPO~\cite{videodpo2024} extends this framework to video diffusion models with omni-preference alignment. In the image domain, preference optimization methods have demonstrated that training on human-preference pairs can substantially improve generation quality along targeted dimensions. However, existing preference optimization approaches use reward signals calibrated for physical realism: structural coherence, photographic quality, and text--image alignment. For anime, this calibration is counterproductive: it penalizes precisely the artistic exaggerations that define the medium---smears and impact frames that exaggerate motion for dramatic effect, or mid-battle shifts to chibi proportions that break tension through tonal contrast---treating them as failures rather than intentional artistry. AniMatrix introduces \emph{deformation-aware preference optimization} with a domain-specific reward model trained on expert-annotated anime preference pairs, enabling the model to learn the distinction between intentional artistry and pathological collapse---a boundary that generic reward models cannot capture.

\subsection{Curriculum Learning for Generative Models}

Curriculum learning~\cite{bengio2009curriculum} organizes training from easy to hard, improving convergence and generalization over na\"{i}ve random sampling. In generative modeling, existing curricula schedule training along axes of \emph{optimization difficulty}---keeping physical realism as a fixed target while varying only sample order: staged resolution or duration schedules (low$\to$high resolution, short$\to$long duration, image$\to$video) are standard in large-scale video DiT systems~\cite{kong2024hunyuanvideo,yang2024cogvideox,wan2025,skyreelsv2}, while noise-difficulty curricula~\cite{kim2024curriculum_diffusion} order denoising tasks by timestep complexity. AniMatrix instead organizes training along \emph{deviation from physics}---style diversity, motion amplitude, and deformation intensity---crossing a heterogeneous spectrum from near-physical to extreme artistic expression, turning the curriculum into both a pacing schedule and a gradual migration of the optimization objective from physical to artistic correctness.

%% file: data_preparation.tex
\paragraph{Production knowledge defines anime correctness.}
We construct the \emph{Production Knowledge System} (PKS) as the first stage of AniMatrix: it redefines ``correct'' anime generation as controllable adherence to production intent. Natural video inherits a shared physical prior, while anime follows many coexisting artistic conventions. PKS turns this production knowledge into training variables through raw acquisition (Sec.~\ref{sec:data:acquisition}), the \emph{Industrial Production Taxonomy} (Sec.~\ref{sec:data:taxonomy}), AniCaption annotation (Sec.~\ref{sec:data:caption}), cascaded curation (Sec.~\ref{sec:data:filtering}), and distribution rebalancing (Sec.~\ref{sec:data:balancing}).

\subsection{Problem Formulation}
\label{sec:data:problem}

In natural video, all training samples share one set of physical laws, so a generative model can converge to a universal implicit prior from any sufficiently large subset of data. Anime provides no such convenience: its training corpus reflects thousands of coexisting artistic conventions---different styles, directors, and schools each following their own internal logic---producing a signal too diverse and contradictory for any single prior to emerge automatically.

Formally, given an unlabeled anime corpus $\mathcal{X}=\{x_1,\dots,x_N\}$, our goal is to construct:
\begin{enumerate}
\item a structured production-variable space $\mathcal{T}=\mathcal{S}\times\mathcal{M}\times\mathcal{C}\times\mathcal{V}$ that decomposes anime's diversity into interpretable, controllable dimensions;
\item a labeled training set $\{(x_i,\,t_i,\,s_i)\}_{i=1}^{n}$ where $t_i\in\mathcal{T}$ is a structured production label and $s_i$ is a natural-language \emph{directorial directive} that encodes directorial intent;
\end{enumerate}
such that the resulting supervision enables the model to learn diverse artistic conventions in an organized, controllable manner.

The distinction from conventional structured captioning is not whether the annotation has a schema, but what kind of variables the schema represents. Recent video-generation systems use structured or dense captions to improve descriptive coverage, decomposing a video into observable entities, actions, background, lighting, camera, style, and atmosphere~\cite{kong2024hunyuanvideo}. Such variables improve prompt following and reconstruction, but they remain attributes of the observed result: they answer what appears in the video. Anime generation requires a different factorization. An anime clip is not merely an observation of a physical process; it is the outcome of production choices. We therefore define $\mathcal{T}$ as a space of \emph{controllable production variables}---style dialect, motion performance, cinematographic choreography, and VFX language---so that conditioning on $t_i$ specifies how a clip should be authored, not merely what content it should contain.

Under this view, AniCaption is not simply a higher-quality structured captioner. It is an inference mechanism from pixels to production decisions: given an observed clip, it estimates its coordinate $t_i$ in $\mathcal{T}$ and verbalizes that coordinate as a natural-language directorial directive $s_i$. Under conditional likelihood training, the target is therefore not only to match a descriptive caption, but to realize a specified production plan. This is the sense in which the PKS redefines correctness: correctness is measured by whether the generated video follows the intended production decisions, rather than whether it merely depicts the same visible content under a physically plausible prior.

We build the PKS in four stages: (i)~acquire a broad, legally compliant raw corpus (Sec.~\ref{sec:data:acquisition}); (ii)~define the Industrial Production Taxonomy and annotate data against it (Sec.~\ref{sec:data:taxonomy} and Sec.~\ref{sec:data:caption}); (iii)~curate data quality through cascaded filtering (Sec.~\ref{sec:data:filtering}); and (iv)~rebalance data coverage (Sec.~\ref{sec:data:balancing}). The taxonomy makes the production variables expressible; AniCaption makes them supervisable; filtering and rebalancing make them learnable at scale.

%% ================================================================
\subsection{Data Acquisition}
\label{sec:data:acquisition}

Unlike general-purpose video generation systems such as HunyuanVideo~\cite{kong2024hunyuanvideo}, which draw from broad multi-domain pools (people, animals, landscapes, vehicles, etc.), AniMatrix targets a single but internally diverse vertical: animated content. This focus demands both \emph{depth}---covering the full spectrum of anime artistic conventions---and \emph{breadth}, spanning decades of production history and a wide range of stylistic paradigms. Depth and breadth are prerequisites for the \emph{Industrial Production Taxonomy} in Sec.~\ref{sec:data:taxonomy}: every production-variable coordinate in $\mathcal{T}$ that never appears in the raw pool is a controllable decision the model cannot learn, regardless of model capacity. Our raw corpus is assembled from publicly available anime media and processed solely for research purposes.

\paragraph{Domain coverage.}
To capture the breadth of anime's artistic landscape, we explicitly balance the raw pool across three diversity axes:
\begin{itemize}
    \item \emph{Rendering paradigm}: traditional cel animation, digital 2D (Retas/Clip Studio Paint workflows), 2D--3D hybrid compositing, and full 3D CG anime---this axis underpins the Style subspace of $\mathcal{T}$ and ensures all major rendering traditions are present before they are discretized into controllable visual-style variables.
    \item \emph{Era}: classic works (pre-2000) through contemporary seasonal anime, ensuring exposure to both hand-drawn traditions and modern digital pipelines so that one production era (e.g., current seasonal digital) does not dominate the implicit data distribution.
    \item \emph{Genre and content}: action/combat, slice-of-life, fantasy, science fiction, mecha, sports, and horror---each genre exercises different subsets of the artistic-exaggeration vocabulary (e.g., combat-heavy \emph{sakuga} vs.\ chibi gags in comedy vs.\ subtle acting in slice-of-life), improving coverage of the Motion and VFX control subspaces in $\mathcal{T}$.
\end{itemize}

\paragraph{Raw pool statistics.}
After basic ingestion---decoding, deinterlacing where necessary, and uniform re-encoding to a lossless intermediate format---we apply coarse entry thresholds: minimum spatial resolution of $480 \times 270$, minimum clip duration of 2\,s, and minimum encoding bitrate to exclude heavily compressed web re-uploads. The cleaned raw pool feeds the subsequent segmentation stage, which produces approximately 150M initial clip segments that are fed to the cascaded curation pipeline of Sec.~\ref{sec:data:filtering}.

%% ================================================================
\subsection{Tag System: Industrial Production Taxonomy}
\label{sec:data:taxonomy}

\paragraph{Design rationale.}
\label{sec:data:taxonomy:rationale}

The purpose of the taxonomy is not to make captions longer or more detailed, but to change the type of variables exposed to the generator. Conventional free-form or structured captions provide an \emph{observation schema}: they decompose a completed clip into visible attributes such as subjects, actions, background, lighting, camera, and style. Anime generation requires a \emph{production-control schema}: variables that specify which creative choices should be controlled to author that result. A caption like ``a girl jumps'' is accurate but erases the choices that make the clip \emph{anime}: anticipation before takeoff, speed lines instead of motion blur, stylized camera follow-through, and impact timing on landing.

Professional anime production naturally provides these variables. Its core stages map onto four orthogonal dimensions of directorial decision-making~\cite{thomas1981illusion}: \emph{art direction} sets visual style and rendering; \emph{performance design} governs motion, acting, and exaggeration; \emph{cinematography} controls framing and camera choreography; and \emph{post-production} adds compositing and VFX. We formalize these dimensions as the \emph{Industrial Production Taxonomy}:
\begin{equation}
    \mathcal{T}=\mathcal{S}\times\mathcal{M}\times\mathcal{C}\times\mathcal{V},
    \label{eq:industrial_taxonomy}
\end{equation}
where $\mathcal{S}$, $\mathcal{M}$, $\mathcal{C}$, and $\mathcal{V}$ denote Style, Motion, Camera, and VFX respectively. The axes are orthogonal in practice: visual identity does not dictate kinetic dialect, so rare combinations such as ``Shinkai Style $\times$ 2D Combat'' remain professionally coherent.

Fig.~\ref{fig:taxonomy_overview} summarizes the full control space and gives a concrete example mapping. Full axis definitions and tag vocabularies are provided in Appendix~\ref{app:taxonomy} (especially Appendix~\ref{app:taxonomy:definitions}). Each tag value is accompanied by a synonym dictionary that maps semantically equivalent aliases (e.g., ``close-up'' $\leftrightarrow$ ``CU'', ``dolly zoom'' $\leftrightarrow$ ``vertigo shot'') to a canonical form; these aliases are used for synonym augmentation during training (Sec.~\ref{sec:model:robustness}).

\definecolor{StyleBlue}{RGB}{168, 197, 224}
\definecolor{StyleBlueLight}{RGB}{230, 240, 250}
\definecolor{StyleBlueDark}{RGB}{60, 110, 165}
\definecolor{MotionGreen}{RGB}{170, 200, 150}
\definecolor{MotionGreenLight}{RGB}{235, 245, 225}
\definecolor{MotionGreenDark}{RGB}{75, 130, 70}
\definecolor{CameraOrange}{RGB}{235, 180, 120}
\definecolor{CameraOrangeLight}{RGB}{252, 240, 220}
\definecolor{CameraOrangeDark}{RGB}{180, 110, 40}
\definecolor{VFXRed}{RGB}{230, 160, 155}
\definecolor{VFXRedLight}{RGB}{252, 232, 230}
\definecolor{VFXRedDark}{RGB}{170, 60, 60}

\begin{figure}[t]
\centering
\begin{tikzpicture}[
  every node/.style={font=\small},
  banner/.style={
    rectangle, rounded corners=6pt,
    draw=black!50, line width=0.8pt,
    fill=black!5,
    minimum width=15.5cm, minimum height=0.95cm,
    align=center, font=\bfseries,
    inner sep=6pt,
  },
  axisbase/.style={
    rectangle, rounded corners=4pt,
    line width=0.7pt,
    minimum width=7.5cm, minimum height=4.0cm,
    text width=6.9cm, align=left, anchor=north west,
    inner sep=8pt, font=\small,
  },
  examplebox/.style={
    rectangle, rounded corners=4pt,
    draw=black!30, line width=0.5pt,
    fill=black!2,
    minimum width=15.5cm,
    text width=14.8cm,
    align=left, font=\footnotesize,
    inner sep=8pt,
  },
]

% --- Top banner ---
\node[banner] (banner) at (0,0)
  {Industrial Production Taxonomy~~$\mathcal{T} = \mathcal{S} \times \mathcal{M} \times \mathcal{C} \times \mathcal{V}$};

% --- Four axis boxes (2x2) ---
\node[axisbase, draw=StyleBlueDark, fill=StyleBlueLight,
      below=0.3cm of banner.south west, anchor=north west] (S) {%
  {\bfseries\large\color{StyleBlueDark} $\mathcal{S}$~~Style}\\[2pt]
  \emph{Visual identity: rendering tradition + kinetic dialect}\\[3pt]
  $\bullet$~\textbf{Visual Style} (11 tags)\\
  \quad 2D: JP / Western / CN / Flat\\
  \quad 3D: Cartoon / Pixar-Disney / Realistic CG\\
  \quad Signature: Miyazaki, Shinkai\\
  $\bullet$~\textbf{Motion Style} (9 tags)\\
  \quad 2D Daily / Combat / Fantasy / \dots\\
  \quad 3D Realistic / Cartoon / Cel-shaded%
};

\node[axisbase, draw=MotionGreenDark, fill=MotionGreenLight,
      right=0.4cm of S.north east, anchor=north west] (M) {%
  {\bfseries\large\color{MotionGreenDark} $\mathcal{M}$~~Motion}\\[2pt]
  \emph{What happens: action + emotion + intensity}\\[3pt]
  $\bullet$~\textbf{Type} (7 categories)\\
  \quad Basic / Movement / Manipulation /\\
  \quad Interaction / Sports / Dance / Vehicle\\
  $\bullet$~\textbf{Emotion} (basic + complex)\\
  \quad happy, anger, fear, melancholy, \dots\\
  $\bullet$~\textbf{Amplitude / Speed} (low / med / high)%
};

\node[axisbase, draw=CameraOrangeDark, fill=CameraOrangeLight,
      below=0.4cm of S.south west, anchor=north west] (C) {%
  {\bfseries\large\color{CameraOrangeDark} $\mathcal{C}$~~Camera}\\[2pt]
  \emph{Cinematography: framing + movement}\\[3pt]
  $\bullet$~\textbf{Shot Scale} (5 levels)\\
  \quad XCU / CU / MS / FS / LS\\
  $\bullet$~\textbf{Viewing Angle} (5 types)\\
  \quad Low / Eye / High / Dutch / Bird's Eye\\
  $\bullet$~\textbf{Movement} (12 types, temporal sequence)\\
  \quad Static, Tracking, Push/Pull, Pan/Tilt,\\
  \quad Orbit, Dolly Zoom, Rack Focus, \dots%
};

\node[axisbase, draw=VFXRedDark, fill=VFXRedLight,
      right=0.4cm of C.north east, anchor=north west] (V) {%
  {\bfseries\large\color{VFXRedDark} $\mathcal{V}$~~VFX}\\[2pt]
  \emph{Anime-specific effects: 80+ tags in 7 categories}\\[3pt]
  $\bullet$~Emotional Symbols~(Vein Pop, Sweat, \dots)\\
  $\bullet$~Character Performance~(Dark Face, Chibi, \dots)\\
  $\bullet$~Animation Techniques~(Smear, Anticipation)\\
  $\bullet$~Action \& Motion FX~(Speed Lines, Impact)\\
  $\bullet$~Skill \& Energy~(Magic Circle, Aura, Lightning)\\
  $\bullet$~Environmental Atmosphere~(God Rays, Petals)\\
  $\bullet$~Physical \& Destruction~(Explosion, Shatter)%
};

% --- Bottom example box ---
% Anchor below the full bottom row (C and V). Using only C.south west places
% the example too high when V is taller than C, so the box overlaps Camera/VFX
% in PDF and especially in HTML/SVG previews.
\node[inner sep=0pt, outer sep=0pt, fit=(C)(V)] (CVrow) {};
\node[examplebox, below=0.4cm of CVrow.south west, anchor=north west] (ex)
  {\textbf{Example mapping for one clip:}~~
   $\mathcal{S}$:~\textit{Shinkai + 2D Daily},~~
   $\mathcal{M}$:~\textit{standing / melancholy / low}\\[2pt]
   \hspace*{2.25cm}
   $\mathcal{C}$:~\textit{MS / Eye Level / Static},~~
   $\mathcal{V}$:~\textit{God Rays + Fog}};

\end{tikzpicture}
\caption{The Industrial Production Taxonomy $\mathcal{T}=\mathcal{S}\times\mathcal{M}\times\mathcal{C}\times\mathcal{V}$. Every clip is mapped to a coordinate in this four-axis production-variable space---Style (rendering paradigm and motion dialect), Motion (performance semantics and kinetic intensity), Camera (cinematographic framing and choreography), and VFX (anime-specific symbolic and technical effects)---forming a structured, navigable control space that the model cannot self-discover from raw pixels. Appendix~\ref{app:taxonomy} details the full axis definitions and vocabularies; Sec.~\ref{sec:data:caption} shows how AniCaption infers these coordinates from clips and verbalizes them as directorial directives.}
\label{fig:taxonomy_overview}
\end{figure}
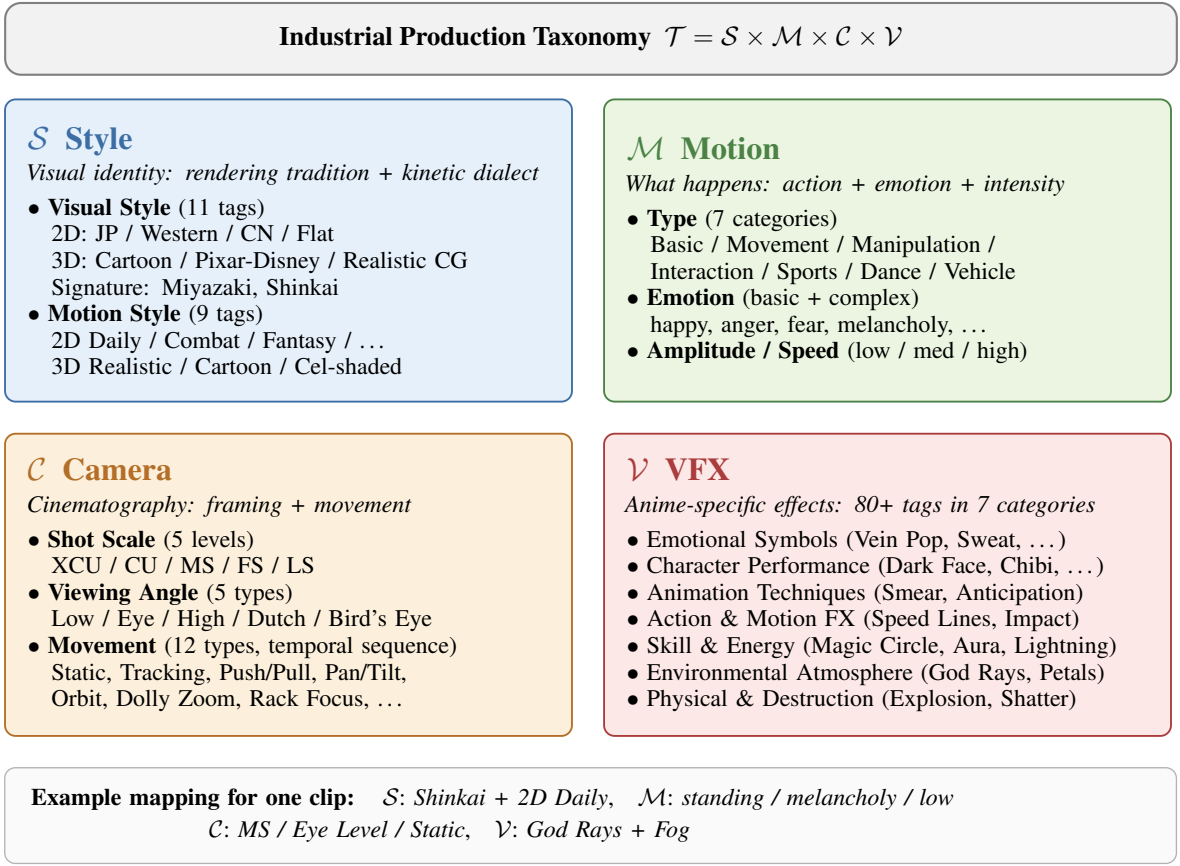

\paragraph{Style axis ($\mathcal{S}$).}
\label{sec:data:taxonomy:style}
Style specifies the authorship mode of a clip: the rendering tradition and kinetic dialect that determine how frames look and how they move. It is represented as the product of \emph{visual style} and \emph{motion style}; full tag definitions are in Appendix~\ref{app:taxonomy:style} (Tables~\ref{tab:visual_style} and~\ref{tab:motion_style}).

\paragraph{Motion axis ($\mathcal{M}$).}
\label{sec:data:taxonomy:motion}
Motion specifies performance semantics: which action occurs, how it is emotionally acted, and how intense its amplitude and speed are. This separates production intent from visible action labels, so running in excitement, running in fear, and running as a combat dash become different motion directives. Full action, emotion, amplitude, and speed definitions are in Appendix~\ref{app:taxonomy:definitions} and Appendix~\ref{app:taxonomy:motion}.

\paragraph{Camera axis ($\mathcal{C}$).}
\label{sec:data:taxonomy:camera}
Camera specifies cinematographic choreography: shot scale, viewing angle, and camera movement, including temporal sequences of camera moves rather than a single global camera tag. This exposes framing and viewpoint as controllable production decisions; full definitions are in Appendix~\ref{app:taxonomy:definitions} and Appendix~\ref{app:taxonomy:camera}.

\paragraph{VFX axis ($\mathcal{V}$).}
\label{sec:data:taxonomy:vfx}
VFX specifies anime's symbolic and technical effects language, from emotion-externalizing symbols to action effects, energy effects, environmental atmosphere, and destruction effects. Each tag carries metadata for meaning, visual appearance, placement/dynamics, and applicable scenes, turning effects into controllable production variables; full definitions and worked examples are in Appendix~\ref{app:taxonomy:definitions} and Appendix~\ref{app:taxonomy:vfx}.

Taken together, the four axes of $\mathcal{T}$ make professional anime \emph{production} knowledge explicit as discrete, composable control variables, supplying a structural role analogous to the universal physical prior that underwrites natural video but that anime data do not provide on their own. Each clip receives a coordinate $t_i\in\mathcal{T}$; AniCaption (Sec.~\ref{sec:data:caption}) estimates this coordinate from pixels and verbalizes it as a natural-language \emph{directorial directive}, while the Tag Encoder (Sec.~\ref{sec:model:tag}) consumes the canonical tags directly. Distribution rebalancing (Sec.~\ref{sec:data:balancing}) then ensures that rare but valid cross-axis combinations remain learnable.

%% ================================================================
\subsection{AniCaption: A Specialized Caption Model for Anime}
\label{sec:data:caption}
\phantomsection\label{sec:data:caption:format}
\phantomsection\label{sec:data:caption:training}
\phantomsection\label{sec:data:caption:eval}

AniCaption is not merely a higher-quality captioner: it is an inference mechanism from pixels to production decisions, estimating the production coordinate $t_i\in\mathcal{T}$ of each clip and verbalizing it as supervision the rest of the system can consume. For every clip it emits two aligned outputs: a fixed-schema structured caption---a JSON object with six field groups (\texttt{subjects}, \texttt{motion}, \texttt{AnimeVisualEffects}, \texttt{style}, \texttt{camera}, \texttt{environment}) whose fields map directly to the taxonomy axes and drive filtering, balancing, and review---and a three-section natural-language directive \texttt{<tag>}/\texttt{<summary>}/\texttt{<description>}, whose canonical taxonomy tags remain machine-readable for the Tag Encoder while the free-form prose supplies the umT5-XXL text channel. Schemas and worked examples are given in Appendix~\ref{app:anicaption:examples}--\ref{app:anicaption:rewrite}.

AniCaption is trained through a four-stage pipeline that progressively raises caption quality: expert sub-models construct an initial structured seed set; Continue-Training (CT) adapts Qwen3-VL to the anime domain and the three-section format on 16M bronze-tier clips; Supervised Fine-Tuning (SFT) uses 500K human-corrected gold-tier clips for production-accurate captions; and Direct Preference Optimization (DPO) targets the two hardest dimensions, motion and VFX (full stage details and Table~\ref{tab:caption_pipeline} in Appendix~\ref{app:anicaption:training}). On a balanced 500-clip held-out set strictly disjoint from training data, AniCaption is evaluated against Gemini~2.5~Pro and Tarsier-2 under both a decompose-then-judge LLM protocol and a human expert protocol by professional anime designers: it attains the best LLM F1 on all three judged dimensions (largest margin on \emph{events}, $+14.0$ over Gemini~2.5~Pro) and the lowest human failure rate on all four dimensions (largest gap on \emph{motion}, 15.4\% vs.\ Gemini's 61.6\%). Full evaluation-set construction, protocols, reliability statistics, per-dimension tables, and figures are in Appendix~\ref{app:anicaption:eval}--\ref{app:anicaption:results} (Table~\ref{tab:caption_eval}).

%% ================================================================
\subsection{Data Curation Pipeline}
\label{sec:data:filtering}
\phantomsection\label{sec:data:cascade}

Data curation must separate technical defects from intentional anime conventions: generic video-quality filters calibrated on live-action footage systematically misread hold frames, limited-animation cadence, smear frames, and stylized compositing artifacts as defects. We therefore apply a coarse-to-fine cascade that first enforces technical validity, then evaluates anime-specific artistic suitability, and finally reserves human expert judgment for high-confidence supervision. General-purpose operators (shot detection, metadata filtering, temporal activity scoring, spatial quality assessment, OCR-based overlay removal, static clip removal, chromatic analysis, and near-duplicate removal) reduce 150M raw clips to 16M technically sound segments. Five anime-specific operators---a learned motion-quality scorer, motion complexity and deformation intensity profilers, and visual style and production era classifiers---further select 6M B-tier clips by evaluating artistic suitability and attaching metadata for rebalancing and curriculum scheduling. Finally, professional anime reviewers evaluate $\geq$4\,s long clips along four axes (motion quality, visual quality, subject coherence, and text--video consistency), admitting 1M clips to A-tier with $>$90\% inter-annotator agreement. A second, stricter expert pass selects from the A-tier pool those clips that professional animators judge fully correct and precise on \emph{every} evaluation dimension, yielding $\sim$500K S-tier clips---the highest-quality subset of the corpus. The three tiers serve different training stages (Sec.~\ref{sec:training}): the 6M B-tier pool, including 2--4\,s clips not promoted, supports continue-training, coverage mixtures, rebalancing, and curriculum bucketing; the A-tier set supplies high-confidence supervision for SFT; and the S-tier subset provides the expert-verified data for Quality Tuning and DPO. Detailed operator definitions, the anime-vs.-live-action comparison (Table~\ref{tab:anime_vs_real}), the full expert rubric, and scorer validation are provided in Appendix~\ref{app:data_curation_section}.

\subsection{Distribution Analysis and Rebalancing}
\label{sec:data:balancing}

Quality filtering ensures individual clip quality but does not guarantee coverage of the full taxonomy space $\mathcal{T}$. Post-curation data exhibits severe long-tail imbalance on every axis: on Motion, dialogue and daily-acting clips dominate ($\sim$45\%) while sakuga and extreme deformation together represent less than 5\%; on Camera, static shots and simple pans account for $\sim$60\% while complex cinematography (dolly zoom, rack focus, dutch angle) appears in fewer than 3\%; on Style, digitally produced anime (post-2010) accounts for $\sim$65\% while classic cel-shaded content comprises only $\sim$8\%. Cross-axis combinations exacerbate the imbalance---a clip exhibiting both close-quarters combat and a dolly zoom in Miyazaki-style rendering may occur fewer than 100 times in the entire corpus. Using the taxonomy labels from AniCaption together with style and era metadata from the anime-specific operators (Appendix~\ref{app:data_curation:anime_ops}), we define a dynamic resampling weight for each clip $x_i$ with label $t_i = (s, m, c, v) \in \mathcal{T}$:
\begin{equation}
    w_i = \left(\frac{1}{n_{s} \cdot n_{m} \cdot n_{c} \cdot n_{v}}\right)^{\alpha},
    \label{eq:resampling}
\end{equation}
where $n_s, n_m, n_c, n_v$ are the marginal counts on each axis and $\alpha \in (0,1)$ controls the degree of flattening ($\alpha{=}0$: original distribution; $\alpha{=}1$: uniform); we set $\alpha = 0.7$ empirically. Beyond frequency rebalancing, we break spurious style--content correlations (e.g., Shinkai-style clips overwhelmingly appear as scenic establishing shots) by constructing a cross-product sampling pool with minimum-representation thresholds per (style, content) pair, preventing shortcuts such as ``Shinkai-style $\Rightarrow$ landscape.''

After rebalancing, the Motion-axis Gini coefficient drops from 0.71 to 0.38, the rarest cross-axis combination grows from fewer than 100 to at least 500 clips, and the A-tier training set comprises approximately 1M clips across three resolution tiers (480p/720p/1080p). Together with taxonomy labels and creator-language directives, this completes the Production Knowledge System: an explicit, structured notion of correctness for anime---in place of the implicit physical prior that natural video inherits for free---exposed to the model through hybrid creator-language conditioning (Sec.~\ref{sec:model:conditioning}) and curriculum scheduling (Sec.~\ref{sec:training}).

%% file: model_design.tex
% ============================================================
%  Section: Model Design
%  AniMatrix — Dual-Channel Creator-Language Conditioned DiT
% ============================================================

AniMatrix adopts the Causal 3D VAE and the Mixture-of-Experts (MoE) DiT backbone from Wan 2.2~\cite{wan2025,peebles2023dit} for latent video generation.
Our architectural contribution is a \emph{dual-channel creator-language conditioning} mechanism that makes the redefined optimization target---artistic correctness rather than physical correctness---architecturally concrete.
We integrate the structured production taxonomy $\mathcal{T}$ constructed in Sec.~\ref{sec:data:taxonomy} directly into the generator through a dedicated \emph{tag encoder} and a \emph{dual-path injection} scheme, enabling simultaneous precise production control and flexible artistic intent.
An overview of the full architecture is shown in Fig.~\ref{fig:architecture}.
We present the conditioning design (Sec.~\ref{sec:model:conditioning}), discuss robustness training and inference control (Sec.~\ref{sec:model:robustness}), and describe production workflow extensions with an efficiency analysis (Sec.~\ref{sec:model:extensions}).

% ------------------------------------------------------------
\subsection{Creator-Language Dual-Channel Conditioning}
\label{sec:model:conditioning}
% ------------------------------------------------------------

Wan 2.2 routes text through a frozen umT5-XXL encoder and injects the resulting token sequence via cross-attention into the DiT~\cite{wan2025,raffel2020t5}.
Our implementation uses the \texttt{models\_t5\_umt5-xxl-enc-bf16.pth} checkpoint with the \texttt{google/umt5-xxl} tokenizer.
This works well for natural video, where a descriptive caption (``a dog runs in a park'') is sufficient.
Professional anime generation, however, requires two fundamentally different types of guidance simultaneously:
\begin{enumerate}[label=(\roman*),nosep]
  \item \emph{Precise production control}---specific, categorical directives that must be followed exactly, such as shot type, camera movement, animation technique, and visual effect (e.g., ``dolly zoom + sakuga combat + speed lines'');
  \item \emph{Flexible artistic intent}---open-ended narrative and stylistic guidance that tolerates interpretation (e.g., ``melancholic farewell under cherry blossoms, Shinkai-style lighting'').
\end{enumerate}
This parallels how a director communicates: the storyboard specifies exact shot composition (non-negotiable), while verbal direction conveys mood and nuance (open to interpretation).
Encoding both into a single text embedding forces the model to guess which parts are strict constraints and which are flexible suggestions---a fundamentally ill-posed task.

\paragraph{Why not tag-augmented text?}
A natural alternative is to prepend serialized tags (e.g., \texttt{``style:cel-2d, motion:sakuga, ...''}) to the free-form text and encode everything jointly through umT5-XXL.
This approach is fundamentally lossy for structured, domain-specific signals, for three reasons:
(1)~the umT5-XXL subword tokenizer fragments domain tags into multiple pieces, so each production tag loses a single learnable identity;
(2)~umT5-XXL self-attention treats the tag sequence as flat text, losing the Cartesian field--value structure of $\mathcal{T}$: the model cannot distinguish that $(\text{style},\,\text{cel-2d})$ and $(\text{motion},\,\text{sakuga})$ belong to orthogonal axes;
(3)~umT5-XXL is frozen to preserve its generalization over open-ended natural language---fine-tuning it on the production vocabulary would improve tag encoding at the cost of degrading free-form text understanding, since structural precision for a closed categorical vocabulary and linguistic flexibility for open-ended narrative are competing objectives within a single encoder.
These observations motivate a dedicated, \emph{trainable} tag encoder that respects the structure of $\mathcal{T}$.

\paragraph{Tag encoder.}
\label{sec:model:tag}
Following the principle of TabTransformer~\cite{huang2020tabtransformer}---which demonstrated that learnable contextual embeddings substantially outperform flat encodings for structured categorical data---we design a lightweight tag encoder that respects the field--value structure of the taxonomy.
Recall that the production taxonomy (Sec.~\ref{sec:data:taxonomy}) is a Cartesian product $\mathcal{T} = \mathcal{S} \times \mathcal{M} \times \mathcal{C} \times \mathcal{V}$, where $\mathcal{S}$, $\mathcal{M}$, $\mathcal{C}$, and $\mathcal{V}$ denote the Style, Motion, Camera, and VFX fields respectively.
Each production tag is a (field, value) pair $(f_i, v_i)$ with $f_i\in\{S,M,C,V\}$; a complete production specification is a set of $k$ tags $\{(f_1,v_1),\dots,(f_k,v_k)\}$ with $k$ typically ranging from 4 to 8.
Each tag is embedded via a compositional decomposition:
\begin{equation}
\label{eq:tag_embed}
e_i = \underbrace{W^{\text{field}}_{f_i}}_{\in\,\mathbb{R}^{d}} + \underbrace{W^{\text{value}}_{v_i}}_{\in\,\mathbb{R}^{d}},
\end{equation}
where $W^{\text{field}}\in\mathbb{R}^{4\times d}$ is a learnable embedding table for the four taxonomy fields and $W^{\text{value}}\in\mathbb{R}^{|\mathcal{V}_{\text{all}}|\times d}$ is a learnable table for the union of all value vocabularies $\mathcal{V}_{\text{all}}=\mathcal{S}\cup\mathcal{M}\cup\mathcal{C}\cup\mathcal{V}$.
The additive decomposition preserves the orthogonal structure of $\mathcal{T}$: tags sharing the same field share the same field embedding, while tags from different fields are distinguished by their field components.

A learnable \texttt{[CLS]} token $e_{\texttt{CLS}}\in\mathbb{R}^d$ is prepended and the full sequence $[e_{\texttt{CLS}},\, e_1,\, \dots,\, e_k]$ is processed by a lightweight Transformer with $N_{\text{tag}}=3$ layers, each comprising multi-head self-attention and a feed-forward network with pre-LayerNorm.
Both embedding tables and all tag encoder parameters are randomly initialized and trained end-to-end with the DiT backbone.
The encoder produces two complementary representations:
\begin{align}
h^{\text{tag}}_{\text{seq}} &= [h_1, \dots, h_k] \in \mathbb{R}^{k \times d},
  &\quad&\text{(per-tag contextualized embeddings)} \label{eq:tag_seq}\\
h^{\text{tag}}_{\text{global}} &= h_{\texttt{CLS}} \in \mathbb{R}^{d},
  &\quad&\text{(global production specification vector)} \label{eq:tag_global}
\end{align}
where $h^{\text{tag}}_{\text{seq}}$ provides fine-grained, per-tag representations for spatially or temporally specific control, and $h^{\text{tag}}_{\text{global}}$ summarizes the entire production specification into a single vector.

\paragraph{Text channel.}
The text channel remains unchanged: the free-form directive $s$ is encoded by the frozen umT5-XXL encoder, yielding a sequence representation $h^{\text{text}}_{\text{seq}} \in \mathbb{R}^{L \times d_{\text{umT5}}}$, where $L$ is the text token count and $d_{\text{umT5}}=4096$.
When $d_{\text{umT5}} \neq d$, a learned linear projection $W^{\text{proj}}\in\mathbb{R}^{d\times d_{\text{umT5}}}$ maps $h^{\text{text}}_{\text{seq}}$ into the shared $d$-dimensional conditioning space before injection.

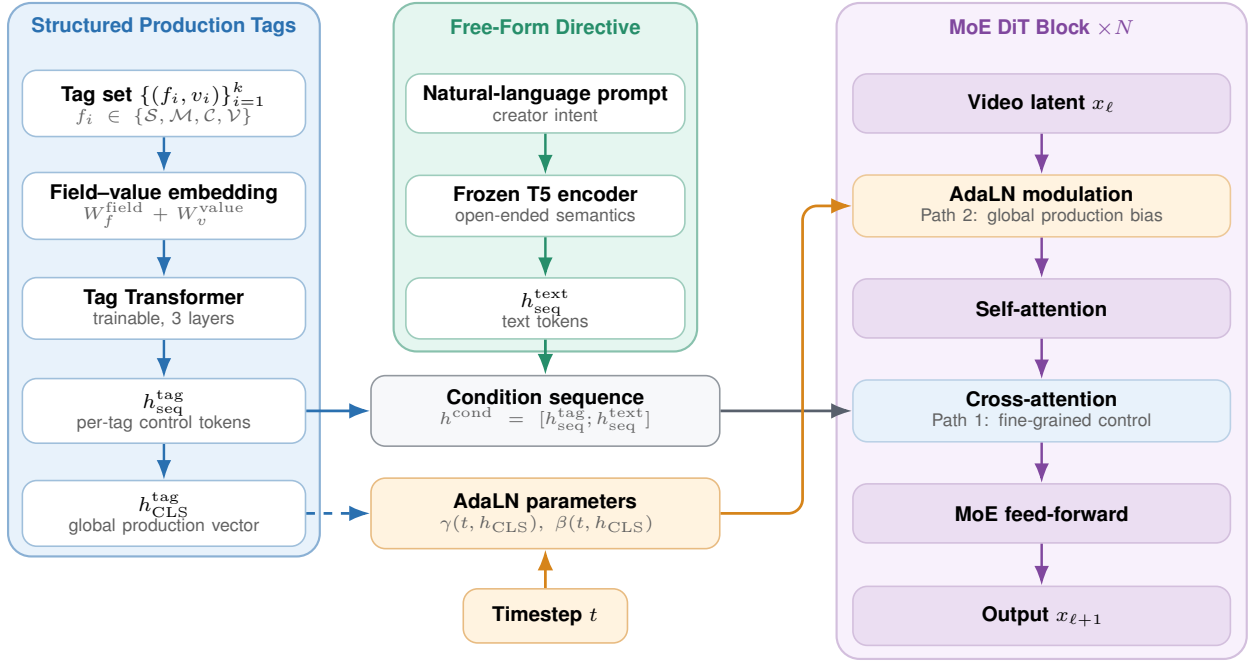
\begin{figure}[t]
\centering
\resizebox{\textwidth}{!}{\input{figures/model_arch.tex}}
\caption{Overview of the Creator-Language Dual-Channel Conditioning architecture. Production tags are encoded by a trainable Tag Transformer via field--value decomposition, while free-form directives pass through a frozen umT5-XXL encoder. The two representations are injected into the MoE DiT through complementary pathways: concatenated sequences via cross-attention (Path~1) for fine-grained spatial/temporal control, and the global tag CLS vector via AdaLN modulation (Path~2) for enforcing overarching production attributes at every layer.}
\label{fig:architecture}
\end{figure}

\paragraph{Path~1: Cross-attention (sequence-level control).}
\label{sec:model:injection}
The two outputs of the tag encoder are injected through complementary pathways into the MoE DiT backbone, inspired by SD3's use of both pooled/global conditioning and token-level conditioning~\cite{esser2024sd3} and grounded in the FiLM conditioning framework~\cite{perez2018film}.
We concatenate the tag and text sequence representations along the token dimension into a unified condition sequence:
\begin{equation}
\label{eq:cond_concat}
h^{\text{cond}} = \left[\,\underbrace{W^{\text{proj}}\,h^{\text{text}}_{\text{seq}}}_{L\;\text{tokens}};\;\underbrace{h^{\text{tag}}_{\text{seq}}}_{k\;\text{tokens}}\,\right] \in \mathbb{R}^{(L+k)\times d},
\end{equation}
and inject it through the MoE DiT's existing cross-attention layers.
This allows each spatial-temporal position in the latent video to attend to both individual tag tokens (for specific production directives) and text tokens (for narrative context).

\paragraph{Path~2: AdaLN modulation (global enforcement).}
Production tags are compact ($k\in[4,8]$), far fewer than free-form text tokens ($L\in[50,100]$).
In a shared cross-attention channel, tag tokens risk being diluted by the numerically dominant text tokens: even under uniform attention, each tag token receives only $\sim\!1/(L{+}k)$ of the total weight.
We additionally inject the global tag representation $h^{\text{tag}}_{\text{global}}$ through a separate \emph{AdaLN modulation} pathway to guarantee enforcement of global production constraints regardless of attention dynamics.

Concretely, at each DiT block $\ell$, the diffusion timestep embedding $t_{\text{emb}}$ and the global tag vector are first fused:
\begin{equation}
\label{eq:adaln_fuse}
c_\ell = \mathrm{SiLU}\!\left(W^{t}_\ell\,t_{\text{emb}} + W^{g}_\ell\,h^{\text{tag}}_{\text{global}}\right),
\end{equation}
where $W^{t}_\ell, W^{g}_\ell \in \mathbb{R}^{d\times d}$ are learnable projections and $\mathrm{SiLU}$ is the activation function~\cite{elfwing2018silu}.
The fused vector $c_\ell$ then produces per-sub-layer AdaLN scale and shift parameters.
Each sub-layer $s\in\{\text{sa},\text{ca},\text{ffn}\}$ (self-attention, cross-attention, FFN) within block $\ell$ maintains its own projection weights:
\begin{equation}
\label{eq:adaln_params}
\gamma_{\ell,s} = W^{\gamma}_{\ell,s}\, c_\ell + \mathbf{1},\qquad
\beta_{\ell,s} = W^{\beta}_{\ell,s}\, c_\ell,
\end{equation}
where $W^{\gamma}_{\ell,s},\,W^{\beta}_{\ell,s}\in\mathbb{R}^{d\times d}$ are per-sub-layer projections; $\gamma_{\ell,s}$ is initialized around $\mathbf{1}$ (via the residual ``$+\mathbf{1}$'') and $\beta_{\ell,s}$ around $\mathbf{0}$ (via zero-initialization of $W^{\beta}_{\ell,s}$), ensuring near-identity modulation at the start of training.
Each sub-layer applies its own modulation as:
\begin{equation}
\label{eq:adaln_apply}
\hat{x} = \gamma_{\ell,s} \odot \mathrm{LayerNorm}(x) + \beta_{\ell,s},
\end{equation}
so that the full block computation becomes:
\begin{equation}
\label{eq:block}
x_{\ell+1} = \mathrm{Block}_\ell\!\left(x_\ell;\;\gamma_\ell,\;\beta_\ell,\;h^{\text{cond}}\right).
\end{equation}
This ensures that overarching production attributes---style, shot grammar, animation technique---are enforced as a global bias on every layer and every sub-layer, independent of attention dynamics.
The dual-path design is summarized in Fig.~\ref{fig:architecture}.

\paragraph{Type embeddings (condition-source disambiguation).}
The dual-channel design separates \emph{how to generate} (production tags vs.\ free-form text).
To prevent source leakage between these semantically distinct directives, we add learnable type embeddings $\tau_{\text{text}},\,\tau_{\text{tag}}\in\mathbb{R}^d$ to the respective condition tokens before cross-attention:
\begin{equation}
\label{eq:type_embed}
\tilde{h}^{\text{cond}} = \left[\,(h^{\text{text}}_i + \tau_{\text{text}});\;(h^{\text{tag}}_j + \tau_{\text{tag}})\,\right].
\end{equation}
This provides the model with an explicit signal to separate structured production tags from free-form creator directives, completing a two-level decoupling: the tag encoder separates structured from free-form conditioning, and dual-path injection separates fine-grained control from global enforcement.

% ------------------------------------------------------------
\subsection{Robustness Training and Inference Control}
\label{sec:model:robustness}
% ------------------------------------------------------------

In real production, the tag and text channels may carry redundant, incomplete, or conflicting information (e.g., tags specify ``close-up'' while text describes a wide landscape).
We design a training strategy that builds robustness to these conditions and exposes explicit control at inference time.

\paragraph{Stochastic conditioning dropout.}
During training, we sample the conditioning mode for each example independently:
\begin{equation}
\label{eq:cond_mode}
m \sim \mathrm{Categorical}\!\left(p_{\text{hybrid}},\;p_{\text{tag}},\;p_{\text{text}},\;p_{\varnothing}\right),
\end{equation}
with $p_{\text{hybrid}}=0.7$, $p_{\text{tag}}=0.1$, $p_{\text{text}}=0.1$, and $p_{\varnothing}=0.1$ (unconditional, for classifier-free guidance~\cite{ho2022cfg}).
Under mode $m=\texttt{tag-only}$, the text tokens are replaced with a learned null embedding $\varnothing_{\text{text}}$; under $m=\texttt{text-only}$, the tag global vector and tag sequence are replaced with corresponding null embeddings $\varnothing^{\text{global}}_{\text{tag}}$ and $\varnothing^{\text{seq}}_{\text{tag}}$; under $m=\varnothing$, both channels are nullified.
This enables \emph{dual classifier-free guidance} at inference.
Following the decomposition principle of Composable Diffusion~\cite{liu2022composable}, we define the tag guidance as the \emph{marginal effect of adding tags given the text condition}, rather than the total effect relative to the unconditional baseline:
\begin{equation}
\label{eq:dual_cfg}
\hat{\epsilon}_\theta = \epsilon_\theta^{\varnothing}
  + \omega_{\text{text}}\!\left(\epsilon_\theta^{\text{text}} - \epsilon_\theta^{\varnothing}\right)
  + \omega_{\text{tag}}\!\left(\epsilon_\theta^{\text{tag+text}} - \epsilon_\theta^{\text{text}}\right),
\end{equation}
where $\omega_{\text{text}}$ and $\omega_{\text{tag}}$ are independent guidance scales.
Increasing $\omega_{\text{tag}}$ strengthens adherence to production tags; increasing $\omega_{\text{text}}$ amplifies narrative fidelity.
Eq.~\eqref{eq:dual_cfg} covers the primary inference mode where both channels are available.
When only tags are provided (no free-form text), the system falls back to standard single-channel CFG: $\hat{\epsilon}_\theta = \epsilon_\theta^{\varnothing} + \omega_{\text{tag}}(\epsilon_\theta^{\text{tag}} - \epsilon_\theta^{\varnothing})$; the tag-only dropout mode ($p_{\text{tag}}=0.1$) ensures $\epsilon_\theta^{\text{tag}}$ is well-calibrated for this scenario.

\paragraph{Partial tag dropping and synonym augmentation.}
Within the \texttt{hybrid} mode, we additionally apply partial tag dropping (each tag is independently dropped with probability $p_{\text{drop}}=0.15$) and synonym substitution (each tag value is replaced with a semantically equivalent alias with probability $p_{\text{syn}}=0.1$, using the synonym dictionary from Sec.~\ref{sec:data:taxonomy}).
Tag dropping teaches the model to infer missing production attributes from context; synonym augmentation reduces overfitting to specific vocabulary.

\paragraph{Controlled tag--text conflict training.}
With probability $p_{\text{conflict}}=0.05$, we inject controlled conflicts between the tag and text channels (e.g., tag specifies ``close-up'' while text is altered to describe a wide shot) and pair the training signal with the \emph{tag-authoritative} ground truth, reinforcing the design intent that tags are hard constraints and text is soft guidance.
This conflict exposure, combined with the dual-CFG mechanism in Eq.~\eqref{eq:dual_cfg}, gives creators explicit, continuous control over the strictness--flexibility trade-off at inference time without retraining.

% ------------------------------------------------------------
\subsection{Production Workflows and Efficiency}
\label{sec:model:extensions}
% ------------------------------------------------------------

\paragraph{Image-to-video (I2V) generation.}
To support I2V workflows where a reference frame specifies character identity and style, we follow Wan 2.2's I2V design~\cite{wan2025} and keep reference-frame conditioning in the VAE latent space.
The reference image $I_0$ is encoded by the VAE to produce $z^{\text{ref}} = E(I_0) \in \mathbb{R}^{1\times h\times w\times c}$, which is concatenated with the noisy latent $z_t$ along the temporal axis as an input-level condition:
\begin{equation}
\label{eq:i2v_input}
\tilde{z}_t = \left[\,z^{\text{ref}};\;z_t\,\right] \in \mathbb{R}^{(1+T')\times h\times w\times c},
\end{equation}
where $T'=T/4$ is the temporally compressed frame count. The VAE latent condition supplies image identity and scene composition, while creator-language controls enter through the text and tag channels.

\paragraph{Spatial super-resolution.}
For production delivery at resolutions above the base model's native output ($512{\times}512$ or $768{\times}768$), we employ a lightweight refinement stage.
The base model generates a mid-resolution latent video $\hat{z}^{\text{LR}}$, which is decoded to pixel space, spatially upsampled via bicubic interpolation, and re-encoded to produce $z^{\text{LR}}_{\uparrow}$.
A second, shallower diffusion model $\epsilon^{\text{SR}}_\phi$ (sharing the same architecture family but with fewer layers) refines this input conditioned on the full conditioning context $\mathcal{G}=\{h^{\text{cond}},\,h^{\text{tag}}_{\text{global}},\,t\}$:
\begin{equation}
\label{eq:sr}
\hat{z}^{\text{HR}} = \text{DDIM-Sample}\!\left(\epsilon^{\text{SR}}_\phi,\;z^{\text{LR}}_{\uparrow},\;\mathcal{G}\right).
\end{equation}
This stage focuses on line sharpness, flat-color cleanliness, and fine detail recovery---qualities critical for anime but less important in natural video---without re-solving the full generation problem.
We train $\epsilon^{\text{SR}}_\phi$ with the same dual-channel conditioning, ensuring production tags are respected throughout the upsampling process.

\paragraph{Trainable vs.\ frozen components.}
The umT5-XXL text encoder and VAE encoder/decoder remain frozen throughout all training stages.
All newly introduced conditioning modules---tag encoder, AdaLN projections ($W^t, W^g, W^\gamma, W^\beta$), text projection $W^{\text{proj}}$, and text/tag type embeddings---are randomly initialized and trained end-to-end with the MoE DiT backbone, which is continue-trained from Wan 2.2~\cite{wan2025}.

%% file: figures/model_arch.tex
\begin{tikzpicture}[
  font=\sffamily\scriptsize,
  >=Latex,
  card/.style={
    rectangle,
    rounded corners=5pt,
    line width=0.55pt,
    minimum height=0.72cm,
    align=center,
    inner xsep=5pt,
    inner ysep=4pt
  },
  group/.style={
    rectangle,
    rounded corners=8pt,
    line width=0.75pt,
    inner sep=7pt
  },
  heading/.style={font=\sffamily\bfseries\scriptsize},
  flow/.style={->, line width=0.85pt, rounded corners=4pt},
  bus/.style={->, line width=0.95pt, rounded corners=5pt}
]

\definecolor{TagBlue}{RGB}{42,112,176}
\definecolor{TagBlueLight}{RGB}{232,242,251}
\definecolor{TextGreen}{RGB}{38,142,106}
\definecolor{TextGreenLight}{RGB}{229,246,240}
\definecolor{FuseGray}{RGB}{88,96,108}
\definecolor{FuseGrayLight}{RGB}{247,248,250}
\definecolor{AdaOrange}{RGB}{214,132,26}
\definecolor{AdaOrangeLight}{RGB}{255,243,224}
\definecolor{DitPurple}{RGB}{128,71,158}
\definecolor{DitPurpleLight}{RGB}{248,241,252}
\definecolor{DitLilac}{RGB}{235,222,242}

% ---------------- Background Boxes ----------------
\node[group, draw=TagBlue!55, fill=TagBlueLight, minimum width=3.80cm, minimum height=6.75cm] (tagbox) at (0,-2.70) {};
\node[heading, text=TagBlue] at ($(tagbox.north)+(0,-0.30)$) {Structured Production Tags};

\node[group, draw=TextGreen!55, fill=TextGreenLight, minimum width=3.70cm, minimum height=4.25cm] (textbox) at (4.65,-1.45) {};
\node[heading, text=TextGreen] at ($(textbox.north)+(0,-0.30)$) {Free-Form Directive};

\node[group, draw=DitPurple!35, fill=DitPurpleLight, minimum width=5.00cm, minimum height=8.00cm] (ditbox) at (10.70,-3.325) {};
\node[heading, text=DitPurple] at ($(ditbox.north)+(0,-0.32)$) {MoE DiT Block $\times N$};

% ---------------- Structured tag channel ----------------
\node[card, draw=TagBlue!45, fill=white, text width=3.10cm] (tags) at (0,-0.55)
  {\textbf{Tag set} $\{(f_i,v_i)\}_{i=1}^{k}$\\[-1pt]
   {\tiny\color{black!62} $f_i\in\{\mathcal{S},\mathcal{M},\mathcal{C},\mathcal{V}\}$}};
\node[card, draw=TagBlue!45, fill=white, text width=3.10cm] (embed) at (0,-1.80)
  {\textbf{Field--value embedding}\\[-1pt]
   {\tiny\color{black!62} $W_f^{\rm field}+W_v^{\rm value}$}};
\node[card, draw=TagBlue!45, fill=white, text width=3.10cm] (tagtr) at (0,-3.05)
  {\textbf{Tag Transformer}\\[-1pt]
   {\tiny\color{black!62} trainable, 3 layers}};
\node[card, draw=TagBlue!45, fill=white, text width=3.10cm] (tagseq) at (0,-4.30)
  {\textbf{$h_{\rm seq}^{\rm tag}$}\\[-1pt]
   {\tiny\color{black!62} per-tag control tokens}};
\node[card, draw=TagBlue!45, fill=white, text width=3.10cm] (tagcls) at (0,-5.55)
  {\textbf{$h_{\rm CLS}^{\rm tag}$}\\[-1pt]
   {\tiny\color{black!62} global production vector}};

\draw[flow, TagBlue] (tags.south) -- (embed.north);
\draw[flow, TagBlue] (embed.south) -- (tagtr.north);
\draw[flow, TagBlue] (tagtr.south) -- (tagseq.north);
\draw[flow, TagBlue] (tagseq.south) -- (tagcls.north);

% ---------------- Free-form text channel ----------------
\node[card, draw=TextGreen!45, fill=white, text width=3.05cm] (prompt) at (4.65,-0.55)
  {\textbf{Natural-language prompt}\\[-1pt]
   {\tiny\color{black!62} creator intent}};
\node[card, draw=TextGreen!45, fill=white, text width=3.05cm] (t5) at (4.65,-1.80)
  {\textbf{Frozen T5 encoder}\\[-1pt]
   {\tiny\color{black!62} open-ended semantics}};
\node[card, draw=TextGreen!45, fill=white, text width=3.05cm] (textseq) at (4.65,-3.05)
  {\textbf{$h_{\rm seq}^{\rm text}$}\\[-1pt]
   {\tiny\color{black!62} text tokens}};

\draw[flow, TextGreen] (prompt.south) -- (t5.north);
\draw[flow, TextGreen] (t5.south) -- (textseq.north);

% ---------------- Fusion nodes ----------------
\node[card, draw=FuseGray!65, fill=FuseGrayLight, text width=3.90cm, minimum height=0.86cm] (concat) at (4.65,-4.30)
  {\textbf{Condition sequence}\\[-1pt]
   {\tiny\color{black!62} $h^{\rm cond}=[h_{\rm seq}^{\rm tag};h_{\rm seq}^{\rm text}]$}};
\node[card, draw=AdaOrange!55, fill=AdaOrangeLight, text width=3.90cm, minimum height=0.86cm] (adalncond) at (4.65,-5.55)
  {\textbf{AdaLN parameters}\\[-1pt]
   {\tiny\color{black!62} $\gamma(t,h_{\rm CLS}),\ \beta(t,h_{\rm CLS})$}};
\node[card, draw=AdaOrange!55, fill=AdaOrangeLight, text width=1.65cm, minimum height=0.66cm] (time) at (4.65,-6.80)
  {\textbf{Timestep} $t$};

\draw[bus, TagBlue] (tagseq.east) -- (concat.west);
\draw[bus, TextGreen] (textseq.south) -- (concat.north);
\draw[bus, TagBlue, dashed] (tagcls.east) -- (adalncond.west);
\draw[bus, AdaOrange] (time.north) -- (adalncond.south);

% ---------------- MoE DiT block ----------------
\node[card, draw=DitPurple!35, fill=DitLilac, text width=4.25cm] (latent) at (10.70,-0.55)
  {\textbf{Video latent} $x_\ell$};
\node[card, draw=AdaOrange!45, fill=AdaOrangeLight, text width=4.25cm] (adaln) at (10.70,-1.80)
  {\textbf{AdaLN modulation}\\[-1pt]
   {\tiny\color{black!62} Path 2: global production bias}};
\node[card, draw=DitPurple!35, fill=DitLilac, text width=4.25cm] (selfattn) at (10.70,-3.05)
  {\textbf{Self-attention}};
\node[card, draw=TagBlue!32, fill=TagBlueLight, text width=4.25cm] (crossattn) at (10.70,-4.30)
  {\textbf{Cross-attention}\\[-1pt]
   {\tiny\color{black!62} Path 1: fine-grained control}};
\node[card, draw=DitPurple!35, fill=DitLilac, text width=4.25cm] (ffn) at (10.70,-5.55)
  {\textbf{MoE feed-forward}};
\node[card, draw=DitPurple!35, fill=DitLilac, text width=4.25cm] (out) at (10.70,-6.80)
  {\textbf{Output} $x_{\ell+1}$};

\draw[flow, DitPurple] (latent.south) -- (adaln.north);
\draw[flow, DitPurple] (adaln.south) -- (selfattn.north);
\draw[flow, DitPurple] (selfattn.south) -- (crossattn.north);
\draw[flow, DitPurple] (crossattn.south) -- (ffn.north);
\draw[flow, DitPurple] (ffn.south) -- (out.north);

\draw[bus, FuseGray] (concat.east) -- (crossattn.west);
\draw[bus, AdaOrange] (adalncond.east) -- ++(1.0,0) |- (adaln.west);

\end{tikzpicture}

%% file: training_strategy.tex
AniMatrix is trained through a four-stage pipeline\footnote{The stage names CT, SFT, and DPO in this section refer to the training of the \emph{AniMatrix video generator}. AniCaption's own CT/SFT/DPO pipeline (Sec.~\ref{sec:data:caption}) trains a separate caption model on different data and objectives.} in which data volume progressively decreases while data quality and optimization specificity increase. Each stage addresses a distinct level of the artistic-correctness objective: (1)~Continue-Training (CT) for large-scale domain adaptation from natural video to anime; (2)~Supervised Fine-Tuning (SFT) for creator-language alignment and progressive curriculum learning; (3)~Quality Tuning (QT) for refinement to professional production standards; and (4)~Deformation-Aware Preference Optimization (DPO) for learning to distinguish intentional artistic exaggeration from structural failure. These four stages instantiate the three-step transition outlined in Sec.~\ref{sec:intro}: CT and SFT jointly \emph{unlearn the physics prior} through domain adaptation and progressive curriculum; QT \emph{refines} the model within the redefined correctness objective; and DPO \emph{distinguishes art from failure} by internalizing a new quality standard. The four stages cannot be merged because adjacent stages differ structurally: CT and SFT operate at different data scales ($\sim$6M broad-coverage vs.\ $\sim$1M precisely labeled) and optimization targets (domain adaptation vs.\ alignment); SFT and QT use different quality tiers (A-tier vs.\ expert-verified S-tier); and QT and DPO employ different supervision types (maximum-likelihood estimation vs.\ pairwise preference).

\subsection{Stage 1: Continue-Training}
\label{sec:training:ct}

The Wan 2.2 foundation model~\cite{wan2025} is pre-trained on natural video, whose motion prior is governed by physics. The goal of continue-training is to shift this prior toward the anime domain at scale, exposing the model to the full breadth of anime styles, motions, and deformations without yet requiring precise control or peak quality.

We initialize training at low resolution and short duration ($256\text{px},\,16\text{f}$) and incrementally scale to target specifications ($720\text{px}+,\,65\text{f}$) across multiple sub-stages, allowing the model to first learn coarse anime semantics (flat colors, exaggerated proportions, non-physical motion patterns) before investing capacity in high-resolution detail and long-duration coherence. To retain the strong object and scene semantics inherited from Wan while shifting the temporal motion prior toward anime, we maintain a balanced mixture of Text-to-Image (T2I), Text-to-Video (T2V), and Image-to-Video (I2V) tasks, treating images as single-frame videos to share the VAE and MoE DiT backbone. The mixture ratio shifts from $\lambda_{\text{T2I}}{:}\lambda_{\text{T2V}}{:}\lambda_{\text{I2V}} = 0.5{:}0.3{:}0.2$ at the lowest resolution sub-stage to $0.2{:}0.4{:}0.4$ at the highest, as temporal capability matures and I2V conditioning becomes a primary production mode. CT uses the broadest data tier---the full B-tier pool of $\sim$6M clips (which subsumes the A-tier and S-tier subsets; see Sec.~\ref{sec:data:filtering})---maximizing coverage of anime's diverse styles and motion types.

\subsection{Stage 2: Supervised Fine-Tuning}
\label{sec:training:sft}

After CT establishes a broad anime prior, SFT has two coupled objectives: (i)~align the model with the creator-language interface (Sec.~\ref{sec:model:conditioning}), and (ii)~progressively expose it to increasingly extreme artistic exaggeration without collapse---addressing the distribution gap identified in Sec.~\ref{sec:intro}.

To align the model with the dual-channel conditioning scheme, we adopt the stochastic conditioning dropout described in Sec.~\ref{sec:model:robustness} (Eq.~\ref{eq:cond_mode}), which jointly trains tag-only, text-only, hybrid, and unconditional modes with partial tag dropping, synonym augmentation, and controlled tag--text conflict exposure. Data in this stage is drawn from A/S-tier clips annotated with AniCaption hybrid prompts (Sec.~\ref{sec:data:taxonomy}).

Directly training on the full spectrum of anime data---including extreme squash-and-stretch, high-speed combat sakuga, and highly diverse style distributions---causes early-training collapse. The root cause is that the model's physics prior and extreme anime exaggeration occupy distant regions of distribution space; forcing the model to bridge this distance in one step is destabilizing. We therefore design a progressive curriculum~\cite{bengio2009curriculum} that provides a \emph{controlled migration path from physical correctness to artistic correctness}, defined along three difficulty axes that each correspond to a distinct way anime departs from physics: \emph{style cluster} $k(x)$ (rendering diversity, from homogeneous to highly varied---departing from photorealistic rendering), \emph{motion amplitude} $m(x)$ (optical-flow energy, from daily acting to combat sakuga---departing from physically plausible kinematics), and \emph{deformation intensity} $d(x)$ (non-rigid flow residuals and keypoint topology stress, from near-physical to extreme artistic deformation---departing from rigid-body geometry). These three axes are directly informed by the motion profiling conducted during data curation (Sec.~\ref{sec:data:filtering}). Each continuous score is discretized into $Q$ quantile buckets: $q_k(x)\in\{1,\dots,Q\}$ for style cluster (assigned by the style classifier of Sec.~\ref{sec:data:filtering}), and $q_m(x),\,q_d(x)\in\{1,\dots,Q\}$ for motion amplitude and deformation intensity respectively. Each clip is then mapped to a difficulty bucket $b(x) = (q_k(x),\, q_m(x),\, q_d(x))$, and the sampling probability follows a sigmoid schedule dependent on training progress $\tau \in [0,1]$:
\begin{equation}
P_\tau(x) \propto w_\tau(b(x)) = \sigma\!\left( \gamma_{\text{cur}} \cdot \left( \tau - \mathcal{D}(b(x)) + \beta_{\text{cur}} \right) \right),
\end{equation}
where $\bar{q}_k=(q_k-1)/(Q-1)$, $\bar{q}_m=(q_m-1)/(Q-1)$, $\bar{q}_d=(q_d-1)/(Q-1)$ are the min-max normalized bucket indices (mapped to $[0,1]$), $\mathcal{D}(b) = \frac{1}{3}(\bar{q}_k + \bar{q}_m + \bar{q}_d)$ is the mean normalized difficulty score, and $\gamma_{\text{cur}}, \beta_{\text{cur}}$ control the curriculum slope and offset. The final per-step sampling probability combines this curriculum weight with the static taxonomy-balancing weight $w_i$ from Sec.~\ref{sec:data:balancing}: $P_\tau(x)\propto w_\tau(b(x))\cdot w_i$, so that both long-tail rebalancing and progressive difficulty scheduling are applied jointly. Early in training ($\tau \approx 0$), simple samples dominate; as $\tau \to 1$, high-deformation, high-style-variance samples are fully introduced. We additionally couple visual difficulty with prompt complexity, concurrently increasing the density of production tags (from $\sim$4 to $\sim$8 tags per clip) and the specificity of text directives as the curriculum introduces visually complex samples.

\subsection{Stage 3: Quality Tuning}
\label{sec:training:qt}

SFT teaches the model \emph{what it can do}; QT refines \emph{how well it does it}. QT uses only S-tier clips---the subset verified by expert reviewers as clean across visual quality, motion coherence, and semantic alignment (Sec.~\ref{sec:data:filtering}). The data proportions are taxonomy-guided: using the labels from AniCaption (Sec.~\ref{sec:data:taxonomy}), we design a target distribution across $\mathcal{T} = \mathcal{S} \times \mathcal{M} \times \mathcal{C} \times \mathcal{V}$ that balances common production scenarios with important but rare combinations (e.g., sakuga combat with complex camera choreography), preventing the model from regressing toward the long-tail head during quality refinement. QT operates at the target production resolution ($720\text{px}+$) and maximum frame count ($65\text{f}$), with a reduced learning rate ($5{\times}10^{-5}$) and shorter training duration focused on polishing motion smoothness, line stability, and color consistency.

\subsection{Stage 4: Deformation-Aware Preference Optimization}
\label{sec:training:dpo}

The final stage completes the transition from physical to artistic correctness by establishing \emph{what counts as right and wrong within the new paradigm}. Once the model generates exaggerated anime motion, both intentional artistic exaggeration and pathological structural failure manifest as ``physics violations,'' making them indistinguishable to standard metrics such as FVD or CLIP score---metrics calibrated for the old, physical-correctness objective. Preference optimization teaches the model to internalize a new quality standard: which deformations are expressive art and which are structural breakdown.

We train a specialized reward model (the ``Judge'') that evaluates generated anime video along four dimensions tailored to anime-specific failure modes: facial topology $r_{\text{face}}$ (consistency of facial geometry under rapid motion), limb structure $r_{\text{limb}}$ (anatomical plausibility during exaggerated poses), line continuity $r_{\text{line}}$ (stability of linework across frames), and motion coherence $r_{\text{motion}}$ (temporal consistency of character identity under deformation). Unlike generic video reward models that penalize any deviation from physical realism, our Judge is trained on anime data and explicitly learns that extreme squash-and-stretch or smear frames are not defects---only structural breakage is. Architecturally, the Judge uses a video encoder initialized from the Wan backbone with four independent linear scoring heads, one per dimension. It is trained on $\sim$20K clips drawn from the A-tier expert-reviewed pool (Sec.~\ref{sec:data:filtering}), where each clip carries human annotations on the four-axis rubric used during data curation (motion quality, visual quality, subject coherence, text--video consistency; see Sec.~\ref{sec:data:filtering}). The Judge distills these holistic human scores into its four anime-specific structural heads ($r_{\text{face}}$, $r_{\text{limb}}$, $r_{\text{line}}$, $r_{\text{motion}}$), each scored on a 1--5 scale. The composite reward is:
\begin{equation}
r(y) = \sum_{j\in\{\text{face},\,\text{limb},\,\text{line},\,\text{motion}\}} w_j\, r_j(y),
\end{equation}
with equal weights $w_j = 0.25$ by default.

Preference pairs $(y_w, y_l)$ are constructed via a semi-automated pipeline. For each of $\sim$10K prompts, the QT-stage model generates $N{=}4$ candidate videos. The Judge assigns composite scores and automatically forms all $\binom{N}{2}$ ordered pairs per prompt where the higher-scored candidate is preferred, rejecting any candidate whose $\min_j r_j$ falls below a threshold of 2.0 (yielding up to 6 pairs per prompt after rejection). For the $\sim$30\% of prompts where the score gap between the top and bottom candidates is small ($\Delta r < 0.5$), expert animators provide additional pairwise preference annotations, yielding $\sim$50K pairs in total. Inter-annotator agreement on the expert-annotated subset exceeds 88\%, consistent with the $>$90\% agreement observed in data curation (Sec.~\ref{sec:data:filtering}). We align the diffusion generator $\pi_\theta$ using Direct Preference Optimization~\cite{rafailov2023dpo,videodpo2024} for its offline stability. The objective maximizes the likelihood margin between preferred and dispreferred outputs:
\begin{equation}
\mathcal{L}_{\text{DPO}}(\theta) = - \mathbb{E}_{(p,\, y_w,\, y_l)} \left[ \log \sigma \left( \beta_{\text{DPO}} \log \frac{\pi_\theta(y_w|p)}{\pi_{\text{ref}}(y_w|p)} - \beta_{\text{DPO}} \log \frac{\pi_\theta(y_l|p)}{\pi_{\text{ref}}(y_l|p)} \right) \right],
\end{equation}
where $\pi_{\text{ref}}$ is a frozen snapshot of the QT-stage model kept fixed throughout DPO, and $\beta_{\text{DPO}}$ controls the deviation penalty.
Because the exact log-likelihood $\log\pi_\theta(y|p)$ is intractable for diffusion models, we follow Wallace et al.~\cite{wallace2024diffusiondpo} and approximate the log-likelihood ratio at each training step by the per-timestep denoising loss difference: $\log\frac{\pi_\theta(y|p)}{\pi_{\text{ref}}(y|p)} \approx -\frac{1}{2}\mathbb{E}_{t,\epsilon}\!\left[\|\epsilon-\epsilon_\theta(z_t,t,p)\|^2 - \|\epsilon-\epsilon_{\text{ref}}(z_t,t,p)\|^2\right]$, where $z_t$ is the noised latent of video $y$ at timestep $t$.

%% file: evaluation.tex
Standard video generation metrics such as FVD~\cite{unterthiner2019fvd} and CLIP score~\cite{radford2021clip} are calibrated for the physical-correctness paradigm and penalize the very artistic exaggerations that define anime, making them unsuitable for evaluating whether a model has successfully transitioned to artistic correctness.
We therefore design an anime-specific human evaluation framework organized around the key quality axes of professional animation production.

\subsection{Experimental Setup}
\label{sec:eval:setup}

We compare AniMatrix against the open-source and closed-source state-of-the-art at the time of evaluation: Wan2.2~\cite{wan2025} (open-source SOTA) and Seedance-Pro~1.0~\cite{seedance2024} (closed-source SOTA marketed for anime). Both baselines are evaluated using their recommended inference configurations. Since these models do not natively support structured production tags, we translate AniMatrix's tag prompts into equivalent natural-language descriptions for fair comparison.

We construct a dedicated test set of 500 prompts covering the full Industrial Production Taxonomy $\mathcal{T}=\mathcal{S}\times\mathcal{M}\times\mathcal{C}\times\mathcal{V}$ (Sec.~\ref{sec:data:taxonomy}). The test set deliberately includes cases that exercise the most challenging regimes of anime generation, such as high-deformation sequences (extreme squash-and-stretch, smears), high-amplitude motion (combat, sakuga), and complex camera choreography (dolly zoom, multi-axis movements), ensuring that the benchmark faithfully reflects the difficulty of real production scenarios. Evaluation is conducted in the image-to-video (I2V) setting: each prompt is paired with a reference first frame that defines subject identity and scene composition, and all five dimensions are scored on the same set of generated outputs.

\paragraph{Evaluation protocol.}
A panel of 15 professional evaluators---each with at least three years of industry experience in animation production or motion graphics---scores every generated clip. Each prompt is independently rated by three evaluators, and the final score is the arithmetic mean. Inter-annotator agreement measured by Krippendorff's $\alpha$ exceeds 0.72 across all dimensions, indicating substantial agreement. We define five anime-specific dimensions, each scored on a 1--5 scale:
\begin{enumerate}[nosep,leftmargin=*]
\item \emph{Style Fidelity}: fidelity of the generated video to the reference first frame in terms of subject identity, scene composition, and overall visual style throughout the clip.
\item \emph{Prompt Understanding}: accuracy with which the video executes the actions, scenes, effects, and other directives specified in the creator-language prompt.
\item \emph{Artistic Motion}: smoothness, naturalness, and rationality of motion, judged by the conventions of the target anime style rather than real-world physics---dramatic exaggerations that serve the narrative (e.g., an impactful blow sending a character flying) are considered correct, while pathological artifacts (e.g., liquid flowing upward, jittery limbs, unnatural freezes) are penalized.
\item \emph{Structural Stability}: freedom from \emph{unintended} visual distortion---warping, tearing, flickering, and morphological breakdown---while explicitly excluding intentional artistic deformations (e.g., squash-and-stretch, smear frames, chibi shifts) that serve directorial intent. This dimension isolates pathological structural failure from the deliberate shape exaggeration scored under Artistic Motion (higher score $=$ fewer genuine artifacts).
\item \emph{Anime Aesthetic}: holistic visual quality including sharpness, detail preservation, clean color gradients, and absence of compression artifacts---a high-quality output should exhibit crisp edges, natural textures, and the polished look of professional anime.
\end{enumerate}

\paragraph{Why not automated metrics?}
We deliberately omit FVD and CLIP score from our main evaluation. In preliminary experiments, we found that these metrics anti-correlate with human quality judgments on anime content: clips with high Artistic Motion scores (strong artistic exaggeration, expressive timing) often receive \emph{worse} FVD scores because their motion departs from the natural-video reference distribution, while static, physics-plausible outputs are rewarded. This confirms that metrics calibrated for physical correctness actively penalize the artistic expression our model is designed to produce. Developing automated metrics that align with professional anime quality judgments remains an important direction for future work (Sec.~\ref{sec:conclusion}).

\subsection{Main Results}
\label{sec:eval:main}

Table~\ref{tab:main_results} reports the human evaluation scores across all five dimensions. AniMatrix ranks first on four of the five dimensions, with near-parity on Structural Stability.

\begin{table}[t]
\centering
\caption{Human evaluation results (5-point scale; higher is better). Best in \textbf{bold}, second-best \underline{underlined}.}
\label{tab:main_results}
\begin{tabular}{lccccc}
\toprule
\textbf{Model} & \textbf{Style Fid.} & \textbf{Prompt Und.} & \textbf{Art.\ Motion} & \textbf{Struct.\ Stab.} & \textbf{Anime Aesth.} \\
\midrule
Seedance-Pro 1.0~\cite{seedance2024} & \underline{4.15} & \underline{3.12} & \underline{3.26} & \textbf{3.84} & \underline{4.09} \\
Wan2.2~\cite{wan2025}          & 4.05 & 2.93 & 3.05 & 3.44 & 3.98 \\
\midrule
\textbf{AniMatrix (Ours)}      & \textbf{4.39} & \textbf{3.82} & \textbf{3.81} & \underline{3.82} & \textbf{4.19} \\
\bottomrule
\end{tabular}
\end{table}

\paragraph{Artistic-expression leadership.}
AniMatrix leads the strongest baseline Seedance-Pro 1.0 by the largest margins on the two dimensions most closely tied to artistic expression and directorial control: \emph{Prompt Understanding} ($+$0.70, $+$22.4\%) and \emph{Artistic Motion} ($+$0.55, $+$16.9\%). This confirms that the dual-channel creator-language conditioning and the style--motion--deformation curriculum---together the two mechanisms most responsible for the transition from physical to artistic correctness---translate directly into measurable gains on the dimensions that define professional anime.

\paragraph{Foundational-dimension saturation.}
On the more foundational dimensions (Style Fidelity, Anime Aesthetic, Structural Stability), differences across models are compressed: the score range across all three models is only 0.21--0.40 on these dimensions, versus 0.76--0.89 on Prompt Understanding and Artistic Motion. This saturation arises because high-tier anime data and competitive base models already yield reasonable rendering and per-frame coherence. AniMatrix still leads on Style Fidelity ($+$0.24) and Anime Aesthetic ($+$0.10), and stays on par with the best baseline on Structural Stability (3.82 vs.\ 3.84, $-$0.02).

\paragraph{Physics-trained models underperform.}
Wan2.2---despite being the strongest open-source baseline in general video generation---scores lowest on Prompt Understanding (2.93) and Structural Stability (3.44) in this anime-specific benchmark. This illustrates that strength on natural video does not transfer to anime and reinforces the need for a domain-specific design.

\subsection{Qualitative Analysis}
\label{sec:eval:qualitative}

\begin{figure}[tbp]
\centering
\includegraphics[width=\textwidth]{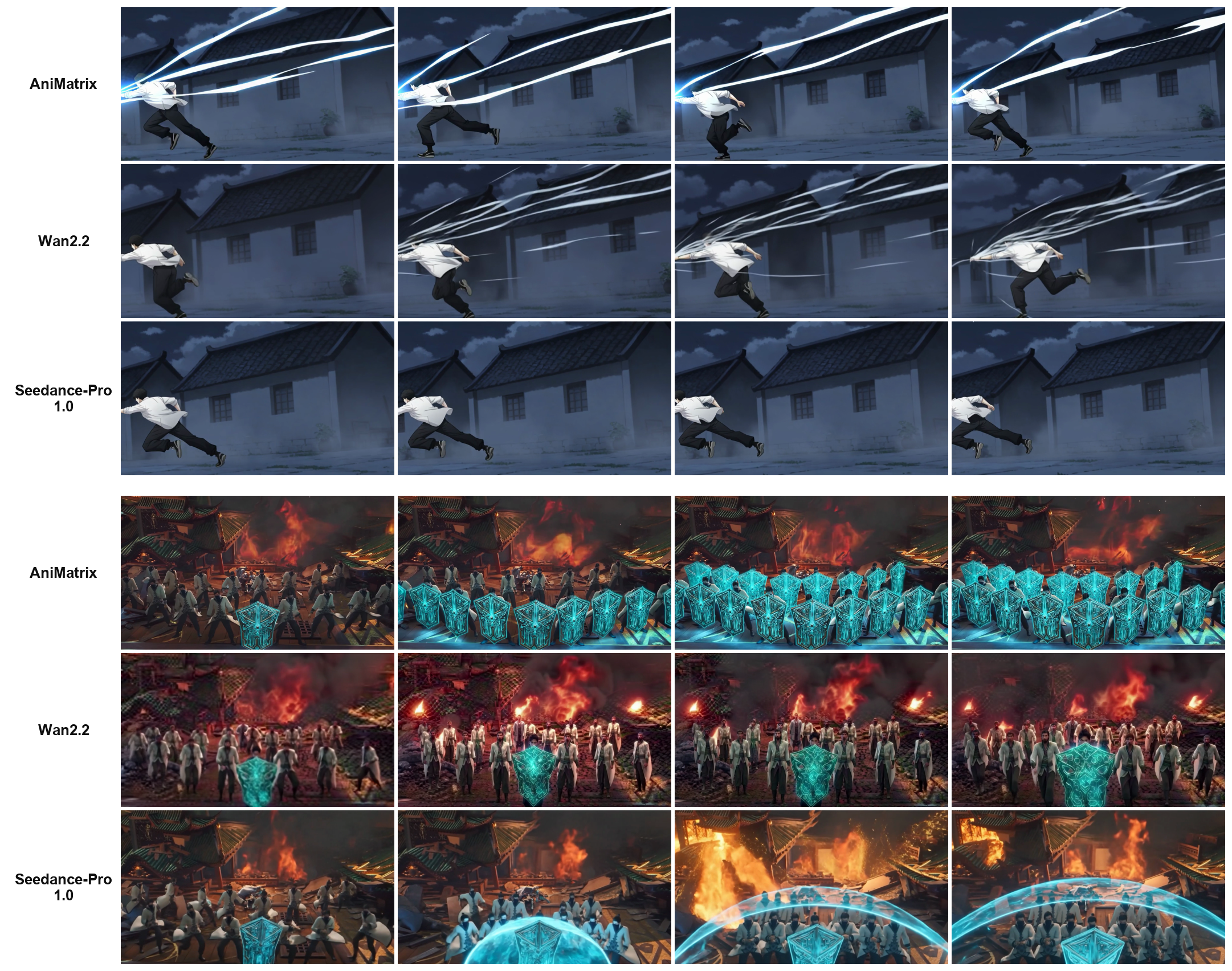}
\caption{Qualitative comparison on two prompts at opposite extremes of the artistic-control spectrum (rows: AniMatrix, Wan2.2, Seedance-Pro 1.0; columns: temporally ordered samples).
\textbf{\emph{Example 1 (top, sakuga).}} A character lunges forward in a low stance, trailed by straight energy beams across the night sky. AniMatrix renders the lunge with crisp straight beams; Wan2.2 collapses the beams into deformed smears with motion blur; Seedance-Pro 1.0 emits no VFX and loses the actor at $t{\approx}1.5$\,s, so its columns are sampled inside the on-screen window.
\textbf{\emph{Example 2 (bottom, group formation with magic shielding).}} Several ancient-costume characters gather into two rows inside a burning building while a blue magic shield condenses and expands, fireballs erupt from the windows, and the camera pushes forward. AniMatrix keeps the group choreography, shield expansion, and fireball timing aligned with the prompt; Wan2.2 forms a looser crowd and localizes the shield; Seedance-Pro 1.0 enlarges the shield but loses formation precision and fireball timing. The baselines fail in distinct artistic-correctness modes (VFX deformation/absence, loose formation, shield localization, timing drift) tracing back to a physics-biased prior.}
\label{fig:qualitative_compare}
\end{figure}

\paragraph{Comparison with Baselines.}
Figure~\ref{fig:qualitative_compare} compares AniMatrix against the open-source and closed-source state-of-the-art at the time of evaluation---\emph{Wan2.2}~\cite{wan2025} (open-source SOTA) and \emph{Seedance-Pro 1.0}~\cite{seedance2024} (closed-source SOTA marketed for anime)---on two prompts at opposite extremes of the artistic-control spectrum.

\emph{Example 1 (high-dynamic sakuga)} couples three controls: character pose (low-stance lunge), VFX form (straight energy beams), and VFX placement (across the night sky behind the character). AniMatrix captures all three. Wan2.2 produces deformed VFX entangled with motion blur, smearing the beams into soft white curls. Seedance-Pro 1.0 captures the lunge but emits no VFX while the actor is on screen; sparks appear only after the character exits at $t{\approx}1.5$\,s.

\emph{Example 2 (group formation with magic shielding)} tests coordinated scene-level control in an epic fantasy setting: several solemn ancient-costume characters gather rapidly into two rows, a blue magic shield condenses and expands, fireballs erupt from the windows of the burning building, and the camera pushes forward slightly. AniMatrix preserves the two-row formation while expanding the shield across the group and sustaining the background fireballs. Wan2.2 forms a looser crowd and concentrates the shield into a smaller foreground mass. Seedance-Pro 1.0 produces a large shield arc, but the group choreography and fireball timing drift from the prompt.

Across both examples (\emph{high-amplitude sakuga} and \emph{coordinated group magic}), AniMatrix satisfies the coupled production directives; the baseline failures---VFX deformation, VFX absence, loose formation, shield localization, and timing drift---trace back to a physics-biased prior treating animation as observed motion over authored production.

%% file: inference_deployment.tex
We accelerate AniMatrix inference $10\times$ via Distribution Matching Distillation (Sec.~\ref{sec:deploy:dmd}) and deploy the resulting system on Workrally at 57-second per-clip latency on an 8$\times$ NVIDIA H20 production node across 60+ anime studios (Sec.~\ref{sec:deploy:scale}).

\subsection{Distribution Matching Distillation for Few-Step Inference}
\label{sec:deploy:dmd}

Distribution Matching Distillation (DMD)~\cite{yin2024dmd} compresses AniMatrix's 40-step I2V Teacher into a Student that runs $10\times$ faster end-to-end, matching or exceeding the Teacher on three of four human-evaluation dimensions while closing to within 0.04 on the fourth (Anime Aesthetic). The 14B MoE DiT uses two noise-level experts; each is distilled with a 4-step DMD schedule, yielding 8 deployment steps in total (Sec.~\ref{sec:deploy:dmd:expert}). We focus the distillation on Image-to-Video (I2V) because anime production starts almost always from a reference frame---a character sheet, a key pose, or the last frame of the previous shot---which the model then animates under a structured production prompt. Unless otherwise stated, every latency and quality number in this section refers to the I2V task. An analogous distillation recipe applies to the T2V variant; each Student specializes in its task-specific conditioning structure.

Distillation touches only the DiT denoiser. The tag encoder, umT5-XXL text encoder, and VAE encoder/decoder reuse the foundation-model weights as-is, so the dual-channel creator-language conditioning of Sec.~\ref{sec:model} carries over by construction. Each task uses three 14B MoE DiT~\cite{wan2025} instances (Table~\ref{tab:dmd_roles}): a trainable Student, a frozen CFG-guided~\cite{ho2022cfg} Teacher, and a Fake/Critic with its own optimizer. Training fits on 64 NVIDIA H800 GPUs with FSDP sharding the three replicas across devices. The Teacher--Student quality comparison reuses the Table~\ref{tab:main_results} dimensions Artistic Motion, Structural Stability, and Anime Aesthetic, and additionally reports Line-Art Quality, a diagnostic subscore for outline continuity, edge crispness, and line flicker.

\begin{table}[H]
\centering
\caption{The three networks in the DMD distillation framework. The Student is trained with a 4-step DMD schedule and deployed with 8 inference steps; the Teacher provides the real-distribution reference; the Fake/Critic tracks the Student's evolving output distribution.}
\label{tab:dmd_roles}
\begin{tabular}{lll}
\toprule
\textbf{Network} & \textbf{Trainable} & \textbf{Role} \\
\midrule
Student        & Yes                       & Denoise at 4 noise levels per expert (8 total at deployment) \\
Teacher        & No (frozen)               & Provide CFG-guided real-distribution reference \\
Fake (Critic)  & Yes (separate optimizer)  & Track Student output for the adversarial signal \\
\bottomrule
\end{tabular}
\end{table}

We use the Flow Matching framework~\cite{lipman2023flow} with linear interpolation $x_t = (1 - \lambda)\, x_0 + \lambda\, \epsilon$, $\lambda \in [0, 1]$, and bias sampling toward high-noise regions via Flow Shift:
\begin{equation}
\lambda_{\text{shifted}} = \frac{s \cdot \lambda}{1 + (s - 1) \cdot \lambda}, \quad s = 10.0.
\end{equation}
Within each expert, we place the four Student anchor steps uniformly across its noise window. For the high-noise expert ($\lambda \in [0.9, 1.0]$), Flow Shift further biases the sampled training noise levels toward $\lambda \to 1$, where denoising decisions determine final video quality and temporal coherence; the low-noise expert ($\lambda \in [0.0, 0.9]$) is distilled analogously (see Per-expert distillation below).

The DMD signal contrasts Teacher and Fake predictions on the same noisy intermediate. Given the Student's predicted clean sample $\hat{x}_0$, we re-noise it to a randomly sampled noise level $\lambda_r$ and form $p_{\text{real}} = \hat{x}_0 - \hat{x}_0^{\text{teacher}}$ and $p_{\text{fake}} = \hat{x}_0 - \hat{x}_0^{\text{fake}}$. The distribution-matching gradient is the normalised difference
\begin{equation}
\text{grad} = \frac{p_{\text{real}} - p_{\text{fake}}}{\left| p_{\text{real}} - p_{\text{fake}} \right|_{\text{mean}} + \epsilon},
\end{equation}
and the Student minimises $\mathcal{L}_{\text{gen}} = \tfrac{1}{2}\big\| \hat{x}_0 - (\hat{x}_0 - \text{grad})_{\text{detach}} \big\|^2$. The loss pushes $\hat{x}_0$ toward $(\hat{x}_0 - \text{grad})_{\text{detach}}$, which approximates the Teacher's clean-sample prediction when the Fake accurately tracks the Student distribution, thereby closing the distributional gap.

We additionally distill dual Classifier-Free Guidance into the Student. The Teacher uses Eq.~\eqref{eq:dual_cfg} with the fixed default pair $\omega_{\text{text}}=5.0$ and $\omega_{\text{tag}}=2.0$ during distillation, combining text guidance for narrative fidelity with tag guidance for production-rule adherence. The Student runs a single forward pass per step at inference (CFG${=}1$): the Teacher's dual-guidance behaviour at this chosen pair collapses into Student weights during distillation. CFG distillation halves per-step compute and removes the runtime CFG knob, multiplying the $5\times$ step-count compression (40 inference steps to 8) by a $2\times$ per-step reduction to deliver $10\times$ wall-clock speedup (Table~\ref{tab:dmd_latency}).
Three mechanisms stabilise adversarial training. \emph{(i)~Asynchronous critic update.} The Fake network updates every two Student steps with a critic-loss-driven adaptive ratio that backs off when Fake loss diverges. \emph{(ii)~Adaptive gradient clip.} An EMA-tracked dynamic threshold $\bar{g} + 3\sqrt{\text{Var}(g)}$ replaces the fixed gradient-norm clip, absorbing scale shifts across noise levels. \emph{(iii)~Student EMA.} An EMA copy of Student weights (decay 0.995) smooths training noise and is the version we deploy.

\paragraph{Per-expert distillation.}
\label{sec:deploy:dmd:expert}
The 14B MoE DiT inherits Wan 2.2's two-expert noise partition~\cite{wan2025}: a high-noise expert covers $\lambda \in [0.9, 1.0]$ (initial denoising from pure noise) and a low-noise expert covers $\lambda \in [0.0, 0.9]$ (refinement). We distill each expert independently with a 4-step DMD schedule (\texttt{num\_student\_timesteps}${=}4$ per expert), preserving the Teacher's expert-routing schedule unchanged. Deployment chains the two 4-step Students into 8 total denoising steps---4 high-noise followed by 4 low-noise---with the same expert boundary at $\lambda{=}0.9$.

\paragraph{Quality--speed takeaway.}
The 8-step Student improves Structural Stability by 0.13 points, Line-Art Quality by 0.08 points, and Artistic Motion by 0.07 points over the Teacher: DMD regularizes away the Teacher's rare deformity outputs that 40-step CFG sampling occasionally produces (Table~\ref{tab:dmd_quality}). The Student trails the Teacher by only 0.04 points on Anime Aesthetic (4.14 vs.\ 4.18), where 40-step CFG sampling preserves slightly finer texture detail. End-to-end I2V latency falls from 577~s to 57~s per clip (Table~\ref{tab:dmd_latency}), a $10\times$ wall-clock speedup that decomposes cleanly into a $5\times$ step-count compression (40 inference steps to 8) and a $2\times$ per-step reduction from CFG distillation.

\begin{table}[H]
\centering
\caption{Quality of the 40-step Teacher vs.\ the deployed 8-step Student (trained with a 4-step DMD schedule) on a held-out I2V evaluation set of 200 anime prompts paired with reference first frames. Scores are 1--5 from professional evaluators. The Student gains 0.13 on Structural Stability, 0.08 on Line-Art Quality, and 0.07 on Artistic Motion by regularizing the Teacher's rare deformity outputs, and trails by 0.04 only on Anime Aesthetic.}
\label{tab:dmd_quality}
\begin{tabular}{lcccc}
\toprule
\textbf{Model} & \textbf{Art.\ Motion} & \textbf{Anime Aesth.} & \textbf{Struct.\ Stab.} & \textbf{Line-Art Qual.} \\
\midrule
40-step Teacher & 4.05 & 4.18 & 3.85 & 4.02 \\
8-step Student  & \textbf{4.12} & 4.14 & \textbf{3.98} & \textbf{4.10} \\
\bottomrule
\end{tabular}
\end{table}

\begin{table}[H]
\centering
\caption{End-to-end I2V latency for one $720{\times}1280$, 5-second clip on an 8$\times$ NVIDIA H20 production node with BF16, FlashAttention-2~\cite{dao2023flashattention2}, and tensor-parallel sharding. The Teacher runs two forward passes per step (CFG); the Student runs one (CFG distilled). The $10\times$ wall-clock speedup decomposes into a $5\times$ step ratio (40 inference steps to 8) and a $2\times$ per-step reduction from CFG distillation.}
\label{tab:dmd_latency}
\begin{tabular}{lccc}
\toprule
\textbf{Model} & \textbf{Steps} & \textbf{Total Latency (s)} & \textbf{Speedup} \\
\midrule
40-step Teacher & 40 & 577         & $1.0\times$ \\
8-step Student  & 8  & \textbf{57} & $\mathbf{10\times}$ \\
\bottomrule
\end{tabular}
\end{table}

\subsection{Deployment at Scale}
\label{sec:deploy:scale}

AniMatrix powers the anime video module of Workrally, Tencent Video's intelligent production platform for full-pipeline anime comic video creation, alongside its text-to-image, image-editing, and audio-generation modules. Three input modes mirror how directors communicate on set. \emph{(i)~Tag panel.} A control panel exposes the Industrial Production Taxonomy (Sec.~\ref{sec:data:taxonomy}) for direct selection of style, motion, camera, and VFX attributes. \emph{(ii)~Prompt box.} A natural-language field accepts both terse instructions and dense creator directives. \emph{(iii)~Reference uploader.} A frame uploader drives I2V workflows. The three modes route into the dual-channel creator-language conditioning of Sec.~\ref{sec:model} and surface its full expressiveness to end users.

\paragraph{Prompt rewriting.}
AniMatrix trains on dense three-section directives (\texttt{<tag>}/\texttt{<summary>}/\texttt{<description>}), but users typically submit brief prompts that omit canonical taxonomy tags. To bridge this distribution gap, the serving stack runs the LLM rewriting pipeline of Appendix~\ref{app:anicaption:rewrite} at inference time: given the user's text---and, when available, the reference frame---the rewriter expands the input into the standardized three-section format. The resulting \texttt{<tag>} fields route to the Tag Encoder as canonical (field, value) pairs, while \texttt{<summary>} and \texttt{<description>} feed the umT5-XXL text channel, so the full dual-channel conditioning surface is exercised even from a single-sentence prompt.

Three deployment numbers translate AniMatrix's benchmark advantage into market outcomes:
\begin{itemize}
\item \emph{Market leadership.} AniMatrix ranks first on download rate inside Workrally's anime business, ahead of Doubao, Kling, and the legacy internal models, among 6 anime-capable systems on the platform.
\item \emph{Inference efficiency.} Online inference latency is 57~s per $720{\times}1280$, 5-second clip on an 8$\times$ NVIDIA H20 production node (Table~\ref{tab:dmd_latency}), driven by the 8-step DMD Student of Sec.~\ref{sec:deploy:dmd} and the serving stack below. We do not report a head-to-head latency table because comparable systems do not publish per-clip latency on matched hardware.
\item \emph{Production scale.} The system serves 60+ studios across 100+ projects.
\end{itemize}

\paragraph{Serving stack.}
The 40-step Teacher uses dual CFG~\cite{ho2022cfg} via Eq.~\eqref{eq:dual_cfg} with $\omega_{\text{text}} \in [4.0, 7.5]$ and $\omega_{\text{tag}} \in [1.0, 3.0]$: lower text scales suit tag-only prompts, higher text scales amplify narrative fidelity for text-heavy directives, and higher tag scales strengthen production-rule adherence. The deployed 8-step Student distills the default pair $\omega_{\text{text}}=5.0$ and $\omega_{\text{tag}}=2.0$ into its weights, so inference exposes no CFG knob. Combined with BF16 inference, FlashAttention-2~\cite{dao2023flashattention2}, and tensor-parallel sharding across the 8$\times$ H20 node, the production stack sustains the 57~s per-clip latency reported in Table~\ref{tab:dmd_latency}.

Three usage patterns recur in production. A \emph{reference-driven} pattern dominates character-centric shots: creators upload a character sheet and specify motion and camera tags, and the model preserves identity while executing the directed action. A \emph{tag-driven} pattern dominates environmental and establishing shots: creators compose the shot from tag combinations alone. A \emph{hybrid} pattern---reference frame plus structured tags plus free-form text---dominates complex narrative shots and exercises the full conditioning surface. Professional creators compose all three modes by shot type, exactly matching the dual-channel design of Sec.~\ref{sec:model}.

The \#1 download share, the 57-second latency, and the 60-studio production footprint together show that AniMatrix's artistic-correctness gains translate into production-line outcomes beyond what benchmark scores capture.

%% file: conclusions.tex
AniMatrix establishes \emph{artistic correctness} as a trainable objective for video generation: artistic motion, stylized appearance, and the expressive shorthand that defines anime. AniMatrix therefore operates on a different axis from current physics-trained video models and world models, which optimize for photorealistic dynamics. Those systems pursue ``looks real,'' while AniMatrix pursues ``looks artistically right.'' We treat artistic correctness as a training signal independent from, and complementary to, physical correctness.

Achieving artistic correctness requires rebuilding both conditioning and supervision, not just swapping the training data. We did this in four steps.
\textbf{(i)~Defining the target.} A Production Knowledge System---a four-axis Industrial Production Taxonomy plus a graph-augmented annotation pipeline (AniCaption)---replaces ``physical-motion fidelity'' with ``directorial-intent fidelity'' as the optimization target.
\textbf{(ii)~Architecting the target.} A dual-channel creator-language conditioning architecture makes this target concrete in the model. A trainable tag encoder preserves the Cartesian field--value structure of the taxonomy. A dual-path injection scheme---cross-attention for fine-grained spatial-temporal control, AdaLN modulation for global enforcement---treats production tags as hard constraints and free-form text as soft guidance, mirroring the storyboard versus verbal-direction split in real anime production.
\textbf{(iii)~Training the artistic prior.} A style--motion--deformation curriculum progressively unlearns the physics prior. A deformation-aware preference optimization stage, paired with a domain-specific reward model, teaches the model to distinguish legitimate artistic violation from failure.
\textbf{(iv)~Production-scale results.} On our anime-specific benchmark AniMatrix ranks first on four of five production dimensions, with near-parity on Structural Stability. The largest gains over Seedance-Pro 1.0 concentrate on Prompt Understanding ($+$0.70, $+$22.4\%) and Artistic Motion ($+$0.55, $+$16.9\%)---gains directly attributable to the dual-channel conditioning architecture and the style--motion--deformation curriculum. In production, AniMatrix runs Workrally at 57\,s per clip on an 8$\times$H20 node, reaches first place in download rate, and serves 60+ studios.

AniMatrix demonstrates that artistic correctness is trainable within a text-conditioned video model, yet the current architecture leaves three structural gaps.
First, conditioning is text-only: character sheets, style references, storyboards, and audio---the assets that drive real anime production---cannot enter the model natively, forcing creators to approximate multimodal intent through language.
Second, artistic motion timing and effect rendering are not first-class conditioning axes; the model still inherits a uniform-motion prior from its physics-pretrained backbone, which dampens the non-uniform rhythms and per-shot effect variation that define professional anime.
Third, generation is a single text-to-video pass with no explicit directorial planning at inference time---shot composition, camera blocking, and scene-level arrangement are left implicit.

We will release \textbf{AniMatrix-Uni}, our next-generation natively multimodal anime generation model, which closes these gaps through three pillars that lift the design from text-to-video to full-pipeline co-creation.
\textbf{(i)~Modality-unified conditioning.} Character sheets, style references, storyboards, and audio (dialogue, music, sound effects) share a single representation. Recent video models---Seedance~2.0~\cite{seedance2}, Sora-2\footnote{Sora-2, Veo-3, and Wan-2.5 are announced products without publicly available technical reports at the time of writing.}, Veo-3, Wan-2.5---push native multimodality on the physical axis; AniMatrix-Uni applies the same architectural shift on the artistic axis, so the conditioning modalities are anime production assets rather than reference photography.
\textbf{(ii)~Artistic motion rhythm and rendering.} Anime motion is non-uniform by design: directors hold, accent, and stretch frames to convey emotion, while effect animation follows a visual grammar that generic motion priors approximate as noise. Seedance~2.0 and Sora-2 optimize for temporally smooth photoreal motion; AniMatrix-Uni instead promotes timing and effect rendering to a first-class artistic conditioning axis. Learning this axis from sakuga sequences and effect-animation supervision will keep high-speed action sharp, land accents on beat, and produce stylized effects with the variability seen in professional production.
\textbf{(iii)~Test-time artistic reasoning.} Before pixel generation, AniMatrix-Uni plans shot composition, camera blocking, and shot-level arrangement under explicit directorial constraints. This mirrors the test-time scaling trend in general video generation, but searches over the Industrial Production Taxonomy rather than physical plausibility---reasoning about directorial intent is the form of test-time compute aligned with artistic correctness.
The aim is a video model that participates in anime creation rather than one that only renders text prompts; the broader claim is that artistic correctness, like physical correctness, scales with native multimodality, motion-and-rendering depth, and test-time reasoning---along an independent axis.

Artistic correctness is a distinct, trainable, and deployable axis of video generation. AniMatrix establishes this axis under production constraints; AniMatrix-Uni will extend it from text-to-video toward fully multimodal anime creation.

%% file: contributors.tex
We gratefully acknowledge all contributors for their dedicated efforts. The following lists recognize participants by their roles. Within the Contributors group, names are listed by surname in Pinyin order. Across all groups, the ordering does not reflect contribution. We also thank the many colleagues from related platform teams whose contributions are not individually listed here.

\begin{itemize}
\item \textbf{Project Sponsors:} Linus, Yu Liu, Qinglin Lu, Yin Zhao
\item \textbf{Project and Tech Lead:} Xiang Wen
\item \textbf{Core Contributors:} ShengJie Wu, Qingwen Gu, Yu Wang, Xin Zheng, Fenghao Zhu, Peng Zhang, HaiTao Zhou, TianXiang Zheng
\item \textbf{Contributors:} Felix Geng, Zhic Gong, Vivi Huang, Reverie Liu, Mina Lu, Yajie Lv, Felix Su, Yifu Sun
\item \textbf{Outside Contributors and Advisors:} Kai Wang
\end{itemize}

%% file: appendix_taxonomy.tex
% Appendix: full vocabularies of the Industrial Production Taxonomy.
% Cross-referenced from Sec.~\ref{sec:data:taxonomy}; main-text mini-summaries
% in Secs.~\ref{sec:data:taxonomy:style}--\ref{sec:data:taxonomy:vfx} cite the
% tables collected here.

The Industrial Production Taxonomy spans 80+ tags across four complementary axes.
The complete tag vocabularies cover all four axes of
$\mathcal{T} = \mathcal{S} \times \mathcal{M}
\times \mathcal{C} \times \mathcal{V}$, defined in Sec.~\ref{sec:data:taxonomy}.
Each subsection corresponds to one axis and lists every tag description in full;
the main paper carries only condensed mini-summaries plus the visual overview
(Fig.~\ref{fig:taxonomy_overview}).

%% ================================================================
\subsubsection{Axis Definitions}
\label{app:taxonomy:definitions}

\paragraph{Style axis ($\mathcal{S}$).}
The Style axis controls the global authorship mode of a clip. Its rendering tradition and kinetic dialect make a clip immediately recognizable as, say, ``Shinkai-style romance'' rather than ``Imaishi-style sakuga combat.'' Style is more than a visual attribute---it is the routing variable that determines the downstream visual budget. A 2D cel-shaded clip and a 3D realistic CG clip differ in outline weight, color palette, shading model, frame-rate convention, and motion timing. Conditioning on a coherent style coordinate therefore constrains all subsequent production choices. Conversely, a mis-specified style coordinate propagates globally, changing the context under which the model learns motion, camera, and VFX.

Unlike natural video, where physical optics determines visual appearance, anime derives its visual identity from deliberate artistic choices along two complementary dimensions. These capture how artists \emph{render} the frames (visual style) and how they \emph{perform} the motion (motion style). A single clip's style is fully specified by the combination of both. The Cartesian product of visual style and motion style produces the full Style space $\mathcal{S}$. Not all combinations are equally common---for instance, ``Shinkai Style $\times$ 2D Combat'' is rare in existing anime but artistically valid---and our distribution rebalancing strategy (Sec.~\ref{sec:data:balancing}) explicitly ensures coverage of such underrepresented but professionally meaningful combinations.

\paragraph{Motion axis ($\mathcal{M}$).}
The Motion axis controls \emph{performance semantics}: not only which action occurs, but how that action is acted, timed, emotionally inflected, and exaggerated. We organize this variable along four complementary sub-dimensions, ordered from categorical semantics to continuous intensity: the categorical \emph{type} of action, the \emph{emotion} conveyed, and two ordinal descriptors---\emph{amplitude} and \emph{speed}---that quantify kinetic intensity. This makes the same visible action controllable under different production meanings: running in excitement, running in fear, and running as a combat dash are different motion directives, not just different captions.

Motion type captures the semantic category of the action being performed under a two-level hierarchy: seven top-level categories, each containing fine-grained action labels. Emotion is annotated as a separate sub-dimension with two levels of granularity, comprising 10 \emph{basic} tags (happiness, anger, fear, etc.) and 10 \emph{complex} tags (melancholy, bittersweet, jealousy, etc.). Motion amplitude measures the spatial extent of movement within the frame, and speed captures the temporal rate of motion. We annotate both as ordinal attributes (\emph{low}, \emph{medium}, \emph{high}); together they determine the dynamic tier of each clip for curriculum scheduling (Sec.~\ref{sec:training:sft}).

\paragraph{Camera axis ($\mathcal{C}$).}
The Camera axis controls cinematographic choreography: how the viewer is positioned relative to the subject, how much of the scene is revealed, and how the viewpoint evolves over time. We decompose this production variable into three controllable sub-dimensions---\emph{shot scale}, \emph{viewing angle}, and \emph{camera movement}. Shot scale spans five levels, viewing angle has five canonical orientations, and camera movement is decomposed into 12 movement types, each (except Static and Shake) annotated with a \emph{direction} and two intensity attributes, \emph{amplitude} and \emph{speed}.

Unlike prior taxonomies that assign a single global camera label per clip, we record \emph{temporal sequences} of camera moves. A single segment chains multiple movements in sequence (e.g., ``static $\to$ medium-speed large-amplitude push-in $\to$ slow pan left''), capturing the compositional grammar of anime cinematography. This sequential representation, combined with per-movement direction and intensity attributes, enables the model to learn complex camera choreographies instead of one opaque label for the whole clip.

\paragraph{VFX axis ($\mathcal{V}$).}
The VFX axis controls the symbolic and technical effects language of anime. Unlike live-action VFX, which often aim to simulate physical phenomena (explosions, weather, particle systems), anime VFX frequently \emph{externalize internal states}: anger becomes a pulsating \textit{Vein Pop} and \textit{Dark Face} mask, embarrassment becomes cascading sweat drops on physics-defying trajectories, surprise becomes a \textit{Statue Pose} petrification gag. The vocabulary---\textit{Smear Frame}, \textit{Anticipation}, \textit{Impact Pose}, \textit{Kira-kira}, and many more---is part of professional craft but is absent from general video-generation training targets.

Our VFX taxonomy captures this professional vocabulary at a granularity that matches industrial practice. Beyond categorical labels, each VFX tag is associated with structured metadata specifying four dimensions: \emph{semantic meaning}, \emph{visual appearance} (shape, color, opacity), \emph{spatial placement and temporal dynamics} (entry, sustain, and exit behavior), and \emph{applicable scenes}. This level of structured metadata is what turns VFX labels into controllable production variables and enables AniCaption (Sec.~\ref{sec:data:caption}) to infer not just \emph{what} effect appears but \emph{how} it should look, move, and behave.

%% ================================================================
\subsubsection{Style Axis Vocabulary}
\label{app:taxonomy:style}

The Style axis $\mathcal{S}$ is decomposed into two sub-axes: \emph{visual
style}, which fixes the rendering tradition (Table~\ref{tab:visual_style}),
and \emph{motion style}, which fixes the kinetic dialect
(Table~\ref{tab:motion_style}). Each clip is assigned exactly one tag along
each sub-axis. See Sec.~\ref{sec:data:taxonomy:style} for the design rationale.

% Table~\ref{tab:visual_style} uses \small font + arraystretch=1.0 so it becomes
% compact enough to share a page with Table~\ref{tab:motion_style}; otherwise
% Table~\ref{tab:visual_style} alone consumes \sim62\%\ of the page and is
% pushed to a page of its own with \sim22\%\ trailing whitespace.
\begin{table}[!htbp]
\centering
\caption{Visual style tags. Each clip is assigned exactly one visual style label.}
\label{tab:visual_style}
\small
\renewcommand{\arraystretch}{1.0}
\begin{tabular}{llp{7.5cm}}
\toprule
\textbf{Group} & \textbf{Tag} & \textbf{Description} \\
\midrule
\multirow{4}{*}{2D}
 & 2D Japanese Anime  & Standard Japanese anime rendering with cel shading, discrete outlines, and characteristic color palettes. \\
 & 2D Western Comics   & Western animation style (e.g., DC/Marvel animated series) with bolder outlines and flatter color blocking. \\
 & 2D Chinese Style    & Chinese ink-wash or Guofeng aesthetics with traditional color schemes and brushstroke-inspired linework. \\
 & 2D Flat Cartoon      & Simplified, flat-design cartoons with minimal shading and geometric character proportions. \\
\midrule
\multirow{3}{*}{3D}
 & 3D Cartoon           & Stylized 3D rendering with exaggerated proportions and cartoon shading. \\
 & 3D Pixar/Disney      & High-quality 3D animation in the Pixar/Disney tradition with subsurface scattering and expressive rigging. \\
 & 3D Realistic CG      & Photorealistic 3D rendering pursuing real-world fidelity in materials, lighting, and physics. \\
\midrule
\multirow{2}{*}{Signature}
 & Miyazaki Style       & Miyazaki's painterly naturalism: lush watercolor backgrounds, hand-drawn fluidity, attention to mundane motion. \\
 & Shinkai Style        & Shinkai's photorealistic lighting composited with stylized characters: lens flares, volumetric light rays, vivid skies. \\
\midrule
 & Live Action          & Real filmed footage (reference plates, rotoscoping sources, or \emph{hybrid} anime+live insert shots); the tag is for \emph{dataset} completeness, not a generative \emph{target} style. \\
 & Other                & Styles not covered above (e.g., pixel art, rotoscoping, mixed media). \\
\bottomrule
\end{tabular}
\end{table}

\begin{table}[!htbp]
\centering
\caption{Motion style tags. Each clip is assigned exactly one motion style label based on its dominant kinetic characteristics.}
\label{tab:motion_style}
\small
% Match Table~\ref{tab:visual_style}'s compact font but use a slightly looser
% \arraystretch so that Tables 6 and 7 expand to fill page 25's tail; Table 5
% keeps the tightest \arraystretch=1.0 since page 24 is already saturated.
\renewcommand{\arraystretch}{1.3}
\begin{tabular}{lp{10cm}}
\toprule
\textbf{Tag} & \textbf{Description} \\
\midrule
2D Daily        & Motion conforms to physical laws; natural daily actions (walking, drinking, gesturing) with correct proportions and perspective. \\
2D Exaggerated  & Exaggerated expressions and body deformations; motion departs from physics (teleportation-like jumps, rubber-hose limbs, ``meme-face'' expression shifts). \\
2D Fantasy       & Magical transformations, sparkle/glow effects, and dreamlike sequences; characteristic of magical-girl and fantasy genres. \\
2D Combat        & Battle choreography with speed lines, impact effects, and dynamic camera work; high kinetic energy and aggressive timing. \\
2D Sci-fi        & Mecha, powered armor, and futuristic environments; motion emphasizes mechanical articulation and technological effects. \\
2D Experimental  & Abstract, non-narrative motion; stream-of-consciousness editing, unconventional transitions, and surreal spatial logic. \\
3D Realistic     & Motion closely mimics real-world physics with high-fidelity material simulation and natural lighting. \\
3D Cartoon       & Stylized 3D motion with cartoon-like squash-and-stretch and exaggerated timing. \\
3D Cel-shaded    & 3D spatial depth and smooth motion combined with 2D-style flat shading, outlines, and hand-drawn aesthetics. \\
\bottomrule
\end{tabular}
\end{table}

%% ================================================================
\subsubsection{Motion Axis Vocabulary}
\label{app:taxonomy:motion}

The Motion axis $\mathcal{M}$ comprises a categorical \emph{type} of action
(Table~\ref{tab:motion_type}), an \emph{emotion} sub-dimension (basic and
complex tags listed below), and two ordinal intensity descriptors
(\emph{amplitude} and \emph{speed}, each in $\{$low, medium, high$\}$). See
Sec.~\ref{sec:data:taxonomy:motion} for the design rationale.

\begin{table}[!htbp]
\centering
\caption{Motion type taxonomy. Seven top-level categories span the full range of anime motion, from basic body mechanics to complex multi-character interaction.}
\label{tab:motion_type}
% Looser \arraystretch lets Table~7 expand to fill the tail of page 25, where
% Table~6 + Table~7 + the A.1.5 lead paragraph would otherwise leave a visible
% strip of trailing whitespace below the A.1.5 lead-in.
\renewcommand{\arraystretch}{1.4}
\begin{tabular}{lp{9.5cm}}
\toprule
\textbf{Category} & \textbf{Representative Actions} \\
\midrule
Basic Actions       & Head: nod, shake, tilt, turn; Face: blink, speak, shout, chew; Torso: bend, bow, twist; Arms: raise, wave, cross; Hands: clap, point, fist; Legs: kick, stomp, tiptoe. \\
Movement \& Posture & Posture change: sit, stand, kneel, lie down, collapse; Horizontal: walk, run, crawl, slide, roll; Vertical: climb, jump, fall; Transitions: sudden stop, sharp turn. \\
Object Manipulation & Carry: lift, hold, drag; Release: throw, drop, toss; Surface: wipe, brush, scrape; Tool use: write, draw, pour, cook, cut, sew. \\
Character Interaction & Friendly: handshake, hug, lean on, link arms; Adversarial: push, tackle, grapple, dodge, strike; Cooperative: support, piggyback, joint carry. \\
Sports \& Athletics  & Ball sports, racket sports, swimming, diving, skating, skiing, martial arts, gymnastics, parkour, extreme sports. \\
Dance \& Performance & Dance, singing, conducting, acrobatics; Instruments: piano, violin, drums, wind instruments. \\
Vehicle \& Driving   & Driving, cycling, piloting; associated mounting/dismounting actions. \\
\bottomrule
\end{tabular}
\end{table}

\paragraph{Emotion tags.}
Emotion is annotated at two levels of granularity, supplementing the motion-type
tag with the affective reading of a clip. The two-tier scheme allows downstream
balancing and conditioning to operate either on the coarse basic tier or on the
finer complex tier:
\begin{itemize}[nosep,leftmargin=*]
\item \emph{Basic emotions}: happiness, excitement, sadness, crying, anger, confusion, disgust, surprise, fear, contempt.
\item \emph{Complex emotions}: shyness, pensiveness, relief, melancholy, bittersweet, tears-of-joy, anxiety, jealousy, pride, gloom.
\end{itemize}

%% ================================================================
\subsubsection{Camera Axis Vocabulary}
\label{app:taxonomy:camera}

The Camera axis $\mathcal{C}$ is decomposed into three sub-dimensions: shot
scale and viewing angle (Table~\ref{tab:camera_frame}), and camera movement
(Table~\ref{tab:camera_movement}, recorded as a temporal sequence of
movements rather than a single global label). See
Sec.~\ref{sec:data:taxonomy:camera} for the design rationale.

\begin{table}[!htbp]
\centering
\caption{Camera framing tags: shot scale and viewing angle.}
\label{tab:camera_frame}
\renewcommand{\arraystretch}{1.5}
\begin{tabular}{llp{8cm}}
\toprule
\textbf{Dimension} & \textbf{Tag} & \textbf{Description} \\
\midrule
\multirow{5}{*}{Shot Scale}
 & Extreme Close-up & Face or object detail filling the frame. \\
 & Close-up         & Character from the chest up, or a localized object view. \\
 & Medium Shot      & Character from the knees up, or partial scene. \\
 & Full Shot        & Complete character figure within a specific scene; the character is the visual center. \\
 & Long Shot        & Character appears small; emphasis on environment, geography, and atmosphere. \\
\midrule
\multirow{5}{*}{Viewing Angle}
 & Low Angle    & Camera tilted upward ($\sim$45\textdegree), looking up at the subject. \\
 & Eye Level    & Horizontal, neutral perspective matching normal human observation. \\
 & High Angle   & Camera tilted downward ($\sim$45\textdegree), looking down at the subject. \\
 & Dutch Angle  & Camera tilted along the roll axis; the horizon appears diagonal, conveying unease or dynamism. \\
 & Bird's Eye   & Camera positioned directly overhead (90\textdegree{} downward), providing a top-down view. \\
\bottomrule
\end{tabular}
\end{table}

\begin{table}[!htbp]
\centering
\caption{Camera movement types. Each movement (except Static and Shake) is further annotated with direction, amplitude, and speed.}
\label{tab:camera_movement}
\renewcommand{\arraystretch}{1.35}
\begin{tabular}{lp{9.5cm}}
\toprule
\textbf{Type} & \textbf{Description} \\
\midrule
Static          & No camera motion. \\
Slight Shake    & Subtle handheld-style tremor for naturalism or tension. \\
Heavy Shake     & Aggressive camera shake for impact, explosion, or disorientation. \\
Tracking        & Camera follows a moving subject, annotated with the subject's position and identity. \\
Push In         & Camera moves toward the subject (dolly in); direction may combine forward with lateral or vertical offset. \\
Pull Out        & Camera moves away from the subject (dolly out); same directional composition as Push In. \\
Translate       & Camera translates laterally or vertically without rotation (e.g., left-to-right, bottom-to-top). \\
Pan / Tilt      & Camera rotates in place: horizontal rotation (pan) or vertical rotation (tilt). \\
Orbit           & Camera revolves around the subject, changing both position and orientation. \\
\midrule
Rack Focus      & Focus shifts between foreground and background subjects within a single shot. \\
Dolly Zoom      & Simultaneous dolly and counter-zoom (Hitchcock/Vertigo effect): subject size remains constant while background scale distorts. \\
\midrule
Other           & Movements not covered above. \\
\bottomrule
\end{tabular}
\end{table}

%% ================================================================
\subsubsection{VFX Axis Vocabulary}
\label{app:taxonomy:vfx}

The VFX axis $\mathcal{V}$ is the largest of the four, with seven top-level
categories and 80+ fine-grained tags (Table~\ref{tab:vfx_tags}). Beyond
categorical labels, each tag carries structured metadata along four
dimensions---\emph{semantic meaning}, \emph{visual appearance}, \emph{spatial
placement and temporal dynamics}, and \emph{applicable scenes}---following the
format of professional anime production sheets;
Table~\ref{tab:vfx_detail} gives representative worked examples. See
Sec.~\ref{sec:data:taxonomy:vfx} for the design rationale.

\begin{table}[!htbp]
\centering
\caption{VFX taxonomy with two-level hierarchy. Seven top-level categories are further divided into subcategories, each containing fine-grained tags drawn from professional anime production terminology. The full taxonomy contains 80+ distinct labels.}
\label{tab:vfx_tags}
\begin{tabular}{llp{10cm}}
\toprule
\textbf{Category} & \textbf{Subcategory} & \textbf{Tags} \\
\midrule
\multirow{6}{*}{\shortstack[l]{Emotional\\Symbols}}
 & Sweat variants      & Single drop, streaming flow, splashing spray---each with distinct emotional intensity (mild unease $\to$ panic). \\
 & Anger marks         & Vein Pop (symbolic cross-shape at temple), nose steam puff, shark teeth / cat fangs. \\
 & Eye transformations & Spiral eyes (dizzy), starry eyes (admiration), heart eyes (infatuation), money-sign eyes, white triangle eyes (menace), dead-fish eyes (apathy). \\
 & Tears variants      & Waterfall tears, single teardrop, corner-eye glistening. \\
 & Floating marks      & Question mark, exclamation mark, lightbulb (idea), musical notes, storm cloud (gloom). \\
 & Other symbols       & Three overhead lines (speechless), blush lines / blush circles, crow gag (awkward silence), nose bubble (sleep). \\
\midrule
\multirow{3}{*}{\shortstack[l]{Character\\Performance}}
 & Face transforms     & Dark Face / shadow mask, charred face (smoke head), melting face, twitching mouth. \\
 & Body transforms     & SD / Chibi form, flat character (paper-thin gag), petrified / statue pose, spiritless (soul leaving body). \\
 & Expression gags     & Electric flash (realization shock), exclamation burst, eye-tail flame (killer intent), saliva spray (shouting). \\
\midrule
\multirow{2}{*}{\shortstack[l]{Animation\\Techniques}}
 & Timing control      & Anticipation, follow-through, slow-in / slow-out, hold pose, snap (instant transition). \\
 & Deformation control & Smear frame, squash \& stretch, overshoot (bounce-back), impact pose, kinetic motion. \\
\midrule
\multirow{3}{*}{\shortstack[l]{Action \&\\Motion Effects}}
 & Speed lines         & Linear (directional), curved (arc motion), radial / converging (burst focus)---each subtype carries different directional semantics. \\
 & Motion traces       & Afterimage / motion blur, motion lines / zip ribbons, smears / exaggeration lines. \\
 & Impact effects      & Impact burst lines, impact flash (white-frame flash), impact frame (inverted-color shock frame), focus lines (radial concentration border). \\
\midrule
\multirow{4}{*}{\shortstack[l]{Skill \&\\Energy Effects}}
 & Light-based         & Light beam / light wave, aura (glowing halo), energy sphere, energy array, highlight glow. \\
 & Electric-based      & Electric shock (current arcs), lightning strike, electric flash (emotion-triggered). \\
 & Fire-based          & Flame trail, fire breath, explosion (energy). \\
 & Magic \& other      & Magic circle (generation / rotation), barrier (shield dome), gunshot muzzle flash. \\
\midrule
\multirow{3}{*}{\shortstack[l]{Environmental\\Atmosphere}}
 & Weather             & Rain (light / heavy / thunder), snow (drift / blizzard), fog / mist, sandstorm, starry night, meteor shower. \\
 & Ambient mood        & God rays (volumetric light), kira-kira sparkles, falling petals / leaves, particles, dream bubbles, emotional background (flowers / hearts for romance). \\
 & Stylized background & Bubble background, gradient light pillar, abstract pattern background, silhouette backdrop. \\
\midrule
\multirow{2}{*}{\shortstack[l]{Physical \&\\Destruction}}
 & Explosion \& debris & Explosion, smoke / dust clouds, shattering / debris scatter, splash (liquid impact). \\
 & Terrain destruction & Ground crack, ground collapse, building collapse, whirlpool / vortex. \\
\bottomrule
\end{tabular}
\end{table}

\begin{table}[!htbp]
\centering
\caption{Detailed annotation examples for representative VFX tags, illustrating the four metadata dimensions specified for each tag in our taxonomy.}
\label{tab:vfx_detail}
\begin{tabular}{l >{\raggedright\arraybackslash}p{2.5cm} >{\raggedright\arraybackslash}p{3.5cm} >{\raggedright\arraybackslash}p{3.8cm} >{\raggedright\arraybackslash}p{2.2cm}}
\toprule
\textbf{Tag} & \textbf{Meaning} & \textbf{Visual Appearance} & \textbf{Placement \& Dynamics} & \textbf{Scenes} \\
\midrule
Vein Pop
 & Anger, irritation, or physical exertion.
 & 3--4 curved ``petal'' shapes forming a cross/clover; bright red or magenta with bold black outline.
 & Appears at temple or forehead; pops in with a bounce, pulses with rhythmic scaling; may stack multiple marks as anger intensifies.
 & Comedic anger, frustration, heated argument. \\
\midrule
Waterfall Tears
 & Exaggerated crying with comedic effect; emotional overflow.
 & Continuous water-curtain bands from both eyes; transparent water-blue with white longitudinal highlights.
 & Pours from lower eyelids down past chin; ``floodgate open'' on emotional trigger; flow volume and speed proportional to intensity.
 & Comedy despair, melodramatic sadness, shock. \\
\midrule
SD / Chibi
 & Comedic deformation; character becomes cute and simplified.
 & Head-to-body ratio shrinks to 1:1--1:3; limbs become stubby cylinders; facial features reduced to dot-eyes and simple mouth symbols.
 & Hard-cut transition from normal proportions; body instantly shrinks while preserving costume colors; reverts on emotional reset.
 & Comedy beats, affectionate moments, chibi interludes. \\
\midrule
Dark Face
 & Suppressed rage, menace, despair, or killing intent.
 & Upper face covered by dark shadow mask (blue-gray to black-purple); eye highlights vanish, replaced by cold-white pinpoint glow.
 & Shadow descends curtain-like from forehead in 1--2 frames; remains static with subtle tremor; lifts on mood shift.
 & Intimidation, villain reveal, emotional breaking point. \\
\midrule
Smear Frame
 & Extreme speed rendered as a single stretched/distorted frame.
 & Character limbs or body elongated along motion trajectory; form becomes a continuous smudge with exaggerated perspective.
 & Inserted for 1--2 frames during peak-velocity moments; surrounded by normal keyframes before and after.
 & Fast punches, whip-pans, rapid head turns, high-speed dashes. \\
\midrule
Speed Lines
 & Visual emphasis on velocity, impact, or focal concentration.
 & Parallel lines (linear), arc-shaped (curved), or converging from edges to center (radial); white or black against contrasting background.
 & Fill background or border region; linear type indicates directional movement, radial type indicates burst or focus; lines flicker with staggered phase.
 & Combat, chase sequences, dramatic reveals, emotional bursts. \\
\midrule
Magic Circle
 & Spell casting, summoning, or power activation.
 & Flat geometric array of concentric rings, runes, and sigils; glowing color (gold, blue, or purple) with additive-blend luminosity.
 & Appears beneath or around caster; generates by tracing outward from center, then rotates continuously; fades or shatters on spell completion.
 & Fantasy combat, transformation, ritual scenes. \\
\midrule
God Rays
 & Warm sunlight streaming through gaps; serene or sacred atmosphere.
 & Diverging trapezoidal light shafts; warm yellow-white with soft gradient edges; semi-transparent overlay.
 & Radiates from upper-frame light source downward; shafts drift slowly with slight shimmer; dust particles float within beams.
 & Morning scenes, forest clearings, emotional resolution, sacred moments. \\
\bottomrule
\end{tabular}
\end{table}

%% file: appendix_anicaption.tex
% Appendix: AniCaption supplementary material.
% Carries the worked examples, full schema description, rewriting details,
% evaluation protocols, and per-dimension evaluation results that the main
% text in Sec.~\ref{sec:data:caption} references in compressed form.

The AniCaption supplement expands the material that the main text
(Sec.~\ref{sec:data:caption}) references in compressed form: complete
worked examples, the full structured-caption schema, the
natural-language rewriting prompt, the LLM-judge and human-expert
evaluation protocols, and per-dimension result tables and figures.

%% ================================================================
\subsubsection{Worked Examples}
\label{app:anicaption:examples}

Both figures below correspond to the same clip---a single still scene
rendered in Shinkai-style with a static medium shot---so that the
structured caption (Fig.~\ref{fig:caption_json_full}) and its
natural-language rewriting (Fig.~\ref{fig:nl_rewrite_full}) can be
compared field-by-field. Compact excerpts are shown first
(Figs.~\ref{fig:caption_json} and~\ref{fig:nl_rewrite}), followed by
their full versions.

\begin{figure}[t]
\centering
\begin{BVerbatim}[fontsize=\small]
"subjects":  [{"idx": 0, "TYPES": {"type": "Human", ...}, ...},
              {"idx": 1, "TYPES": {"type": "Vehicles", ...}, ...}]
"motion":    [{"idx": 0, "action": "<subject_0> remains still ...",
               "expression": "...", "amplitude": "low"}]
"AnimeVisualEffects": {
   "HasAnimeVisualEffects": true,
   "AnimeVisualEffectsStructure":
     [{"TYPES": {"type": "Environmental",
                 "sub_type": "Weather", "sub_sub_type": "Fog"},
       "position": "background", "description": "..."}]}
"VideoStyle": "Shinkai Style", "MotionStyle": "2D Daily",
"shot_type": "medium shot",    "camera_motion": "static",  ...
\end{BVerbatim}
\caption{Compact excerpt of a structured caption highlighting three distinguishing design choices: (i)~the temporally ordered \texttt{motion} array uses cross-references such as \texttt{<subject\_0>} to subjects; (ii)~the \texttt{AnimeVisualEffects} field carries the three-level VFX hierarchy (\texttt{type}/\texttt{sub\_type}/\texttt{sub\_sub\_type}); (iii)~global style and camera tags are kept separate from per-entity annotations.}
\label{fig:caption_json}
\end{figure}

\begin{figure}[t]
\centering
\begin{BVerbatim}[fontsize=\small]
{
  "subjects": [
    { "idx": 0,
      "TYPES": {"type": "Human", "sub_type": "Woman"},
      "appearance": "long platinum blonde hair in two braids, ...",
      "position": "Centrally positioned in the frame." },
    { "idx": 1,
      "TYPES": {"type": "Vehicles", "sub_type": "Ship"},
      "appearance": "dark-colored with multiple masts, ...",
      "position": "In the background, blurred." }
  ],
  "motion": [
    { "idx": 0,
      "action": "<subject_0> remains still, looking forward ...",
      "expression": "neutral, slight melancholy",
      "amplitude": "low" }
  ],
  "AnimeVisualEffects": {
    "HasAnimeVisualEffects": true,
    "AnimeVisualEffectsDescription": "Soft blue glow and ...",
    "AnimeVisualEffectsStructure": [
      { "idx": 0,
        "TYPES": {"type": "Environmental",
                  "sub_type": "Weather", "sub_sub_type": "Fog"},
        "position": "background",
        "description": "Flowing waterfall-like mist with ..." }
    ]
  },
  "MotionAmplitude": "low", "MotionStyle": "2D Daily",
  "VideoStyle": "Shinkai Style",
  "shot_type": "medium shot", "shot_angle": "eye level",
  "camera_motion": "static",
  "environment": "fantasy harbor", "lighting": "soft twilight"
}
\end{BVerbatim}
\caption{Full structured caption for a single clip, expanded from the excerpt in Fig.~\ref{fig:caption_json}. Fields are organized into six groups: \texttt{subjects} (entity identity and position), \texttt{motion} (temporal action and expression), \texttt{AnimeVisualEffects} (hierarchical VFX annotations), global style tags, camera metadata, and \texttt{environment} (scene description). Cross-references such as \texttt{<subject\_0>} link motion descriptions to specific subjects.}
\label{fig:caption_json_full}
\end{figure}

\begin{figure}[t]
\centering
\begin{BVerbatim}[fontsize=\small]
<tag> VideoStyle: Shinkai Style, MotionStyle: 2D Daily,
      shot_type: medium shot, camera_motion: static, ...
<summary> A blonde woman stands still at a fantasy harbor
at twilight, gazing forward with quiet melancholy.
<description> A woman with long platinum blonde hair ...
The camera holds a static medium shot at eye level. She
remains still, ... In the background, a dark multi-masted
ship is docked, rendered in soft blur. ...
\end{BVerbatim}
\caption{Compact excerpt of the three-section natural-language directive rewritten from Fig.~\ref{fig:caption_json}. Machine-readable \texttt{<tag>} keeps the canonical taxonomy form for the Tag Encoder (Sec.~\ref{sec:model:tag}); \texttt{<summary>} and \texttt{<description>} carry human-readable prose for the umT5-XXL text encoder.}
\label{fig:nl_rewrite}
\end{figure}

\begin{figure}[t]
\centering
\begin{BVerbatim}[fontsize=\small]
<tag> VideoStyle: Shinkai Style, MotionStyle: 2D Daily,
MotionAmplitude: low, shot_type: medium shot,
shot_angle: eye level, camera_motion: static

<summary> A blonde woman stands still at a fantasy harbor
at twilight, gazing forward with quiet melancholy.

<description> A woman with long platinum blonde hair styled
in two braids stands centrally in the frame, wearing a blue
cloak. The camera holds a static medium shot at eye level.
She remains still, looking forward with a neutral expression
tinged with slight melancholy. In the background, a dark
multi-masted ship is docked, rendered in soft blur. The
scene is enveloped in flowing waterfall-like mist with a
soft blue glow, and several massive rocks float suspended
in the air. The environment is a fantasy harbor bathed in
soft twilight lighting.
\end{BVerbatim}
\caption{Full natural-language rewriting output for the structured caption in Fig.~\ref{fig:caption_json_full}, expanded from the excerpt in Fig.~\ref{fig:nl_rewrite}. The three-section format (\texttt{<tag>}, \texttt{<summary>}, \texttt{<description>}) separates machine-readable tags from human-readable prose while ensuring all structured fields are faithfully represented.}
\label{fig:nl_rewrite_full}
\end{figure}

%% ================================================================
\subsubsection{Structured Caption Schema}
\label{app:anicaption:format}

Table~\ref{tab:caption_fields} lists the six field groups of the
AniCaption structured caption format introduced in
Sec.~\ref{sec:data:caption:format}, together with the downstream data
engineering function each group serves.

\begin{table}[!htbp]
\centering
\caption{Field groups in the AniCaption structured format. Each group maps to one or more taxonomy axes and serves specific downstream functions.}
\label{tab:caption_fields}
\begin{tabular}{lp{5.5cm}p{4.5cm}}
\toprule
\textbf{Field Group} & \textbf{Contents} & \textbf{Downstream Use} \\
\midrule
\texttt{subjects}
 & Per-subject entries with semantic type (human / animal / vehicle / object), sub-type, visual appearance description, and spatial position in frame.
 & Subject-aware filtering; character consistency analysis; reference-frame conditioning. \\
\midrule
\texttt{motion}
 & Temporal sequence of motion segments, each specifying the acting subject (via cross-reference), action description, facial expression, and motion amplitude.
 & Motion-type classification; deformation profiling; curriculum bucketing. \\
\midrule
\texttt{AnimeVisualEffects}
 & Boolean presence flag; free-text overall description; and a structured array of individual effects, each with hierarchical type labels (type / sub\_type / sub\_sub\_type from the VFX taxonomy), spatial position, and per-effect description.
 & VFX-aware data selection; effect-specific training; tag vocabulary for the Tag Encoder. \\
\midrule
\texttt{style}
 & Global tags: \texttt{VideoStyle}, \texttt{MotionStyle}, \texttt{MotionAmplitude}---drawn directly from the Style axis vocabulary.
 & Style-based filtering and rebalancing; curriculum style-cluster assignment. \\
\midrule
\texttt{camera}
 & Global tags: \texttt{shot\_type}, \texttt{shot\_angle}, \texttt{camera\_motion}, and \texttt{lighting}.
 & Camera-aware selection; cinematography conditioning. \\
\midrule
\texttt{environment}
 & Free-text scene description capturing the setting, location, and spatial context of the clip (e.g., ``fantasy harbor at dusk,'' ``school rooftop under cherry blossoms'').
 & Scene-based retrieval; environment-conditioned generation. \\
\bottomrule
\end{tabular}
\end{table}

Three design choices distinguish this schema. First, the \texttt{motion} array is \emph{temporally ordered} and uses \emph{cross-references} to subjects (e.g., \texttt{<subject\_0>}), enabling multi-subject temporal reasoning---critical for scenes where ``Character~A attacks Character~B while the camera dollies in.'' Second, the \texttt{AnimeVisualEffects} field combines a free-text summary with a structured array of individual effects, each tagged with the three-level VFX taxonomy hierarchy (type / sub\_type / sub\_sub\_type). This dual representation supports both coarse-grained filtering (``does this clip contain any VFX?'') and fine-grained analysis (``how many clips have radial speed lines co-occurring with impact frames?''). Third, the global style and camera tags are deliberately separated from the per-subject and per-effect annotations, reflecting their role as clip-level production directives rather than entity-level descriptions.

%% ================================================================
\subsubsection{Natural-Language Rewriting Details}
\label{app:anicaption:rewrite}

The main-text rewriter (Sec.~\ref{sec:data:caption:format}) is built from
two complementary LLMs that share a single prompt template and structural
schema. We use a strong proprietary model (Claude) for the highest-quality
tier of training data, where fluency and faithful preservation of nuance
matter most; and an open-source model (Qwen3) for the bulk of the corpus,
where comparable structural fidelity at a fraction of the per-call cost is
required to support in-house batch processing at the 16M-clip scale of
Continue-Training (Sec.~\ref{sec:data:caption:training}). Because both
rewriters consume the same structured caption and follow the same prompt
contract, the resulting natural-language directives are stylistically
consistent across the corpus regardless of which rewriter produced them.

The output of either rewriter is the standardized three-section format:

\begin{itemize}[nosep,leftmargin=*]
\item \texttt{<tag>}~Global production tags serialized as key-value pairs: \texttt{VideoStyle}, \texttt{MotionStyle}, \texttt{MotionAmplitude}, \texttt{shot\_type}, \texttt{shot\_angle}, and \texttt{camera\_motion}. These tags are kept in their canonical taxonomy form for direct consumption by the Tag Encoder (Sec.~\ref{sec:model:tag}).
\item \texttt{<summary>}~A single sentence summarizing the overall content of the clip, providing a concise overview for coarse-grained understanding.
\item \texttt{<description>}~A detailed, temporally organized paragraph integrating all structured fields into fluent creator-language prose. The description follows a consistent ordering: \emph{subject appearance and identity} $\to$ \emph{camera framing and movement} $\to$ \emph{character motion and expression} $\to$ \emph{visual effects} $\to$ \emph{scene and environment}, ensuring that fine-grained production details---specific animation techniques, VFX timing, camera choreography---are preserved rather than collapsed into vague summaries.
\end{itemize}

The rewriting prompt instructs the model to (i)~temporally order all events described in the \texttt{motion} array, (ii)~integrate VFX descriptions at the temporal points where they occur, (iii)~convert taxonomy tags into professional anime terminology, and (iv)~maintain strict consistency with the structured source fields. This ensures that every natural-language directive is \emph{grounded in and traceable to} its structured caption, avoiding the semantic drift that would occur if the two formats were produced independently.

%% ================================================================
\subsubsection{Training Details}
\label{app:anicaption:training}

We train AniCaption through a four-stage pipeline that tightly interleaves data construction and model optimization---each stage both produces higher-quality data and yields a stronger model, forming an iterative refinement loop. Table~\ref{tab:caption_pipeline} summarizes the stages; the paragraphs below provide the implementation details summarized in Sec.~\ref{sec:data:caption:training}.

\begin{table}[!htbp]
\centering
\caption{AniCaption training pipeline. Data construction and model training are interleaved across four stages, with each stage producing both improved data and a stronger model.}
\label{tab:caption_pipeline}
\begin{tabularx}{\textwidth}{l X c X}
\toprule
\textbf{Stage} & \textbf{Data} & \textbf{Scale} & \textbf{Objective} \\
\midrule
Expert Models  & Per-dimension annotated anime data            & $\sim$50K  & Train specialized sub-models per taxonomy axis \\
CT             & Bronze-tier anime with model captions         & $\sim$16M  & Domain adaptation to anime distribution and format \\
SFT            & Gold-tier anime with human-corrected captions & $\sim$500K & Precise caption quality through expert supervision \\
DPO            & Preference pairs for motion \& VFX            & ---        & Targeted improvement on motion and VFX descriptions \\
\bottomrule
\end{tabularx}
\end{table}

\paragraph{Stage 1: Expert Sub-Models.}
We build specialized \emph{expert sub-models}, each focused on one dimension of the structured caption. For each taxonomy axis---video style, motion style, camera (shot type, angle, movement), VFX, and others---we collect approximately 50K annotated anime clips with per-dimension labels and train a dedicated classifier or descriptor. We apply these expert models to the full corpus: each sub-model annotates its corresponding dimension. We then assemble the per-dimension outputs into complete structured captions (Sec.~\ref{sec:data:caption:format}). This decompose-then-compose strategy outperforms end-to-end structured prediction, because each expert trains on clean, dimension-specific supervision rather than noisy joint labels. From the assembled structured captions, we select approximately 20K high-confidence samples for human review: annotators verify and correct the composite structured captions, producing a seed set of validated structured annotations.

\paragraph{Stage 2: Continue-Training (CT).}
We convert the validated structured captions to natural-language format via the LLM rewriting pipeline (Sec.~\ref{sec:data:caption:format}). We then perform continue-training on Qwen3-VL using approximately 16M bronze-tier anime clips, each paired with its model-generated natural-language caption. The objective of CT is twofold: (i)~adapt Qwen3-VL's visual representations from its general-domain pretraining distribution to the anime domain, and (ii)~familiarize the model with the three-section output format (\texttt{<tag>}, \texttt{<summary>}, \texttt{<description>}). At this stage, caption quality does not yet reach production grade, but the model acquires the domain-specific visual vocabulary and format conventions needed for subsequent refinement.

\paragraph{Stage 3: Supervised Fine-Tuning (SFT).}
SFT produces the largest quality jump in the pipeline. We fine-tune the CT-adapted model on approximately 500K gold-tier clips, selected from the broader corpus via taxonomy-guided category balancing to ensure proportional representation across styles, motion types, camera techniques, and VFX categories. Each caption in this set undergoes human expert correction: annotators review the model-generated structured captions and natural-language descriptions, correcting factual errors (e.g., misidentified camera movement), adding missing details (e.g., overlooked VFX), and refining the natural-language phrasing to match professional production vocabulary. This transforms the model from generating plausible-but-approximate captions to producing precise, production-accurate annotations.

\paragraph{Stage 4: Preference Optimization (DPO).}
SFT yields strong overall caption quality, but two dimensions benefit from targeted refinement: \emph{motion descriptions} (where subtle differences in timing, amplitude, and deformation are difficult to capture) and \emph{VFX descriptions} (where overlapping effects and their temporal co-occurrence patterns require precise articulation). We construct preference pairs by generating multiple caption candidates for the same clip, then having expert annotators select the preferred description for motion and VFX segments. Direct Preference Optimization~\cite{rafailov2023dpo} is applied to align the model toward more accurate and detailed motion and VFX descriptions without degrading performance on other dimensions.

%% ================================================================
\subsubsection{Evaluation Protocols}
\label{app:anicaption:eval}

The evaluation specification covers the two protocols summarized in
Sec.~\ref{sec:data:caption:eval}: the
LLM-as-a-judge protocol and the human-expert protocol. Both run on the
same 500-clip held-out evaluation set described in
Sec.~\ref{sec:data:caption:eval}.

\paragraph{Evaluation set.}
We construct a held-out evaluation set of 500 anime clips that is strictly disjoint from all training data used in any stage (Sec.~\ref{sec:data:caption:training}). To prevent biased conclusions driven by skewed sampling, the set is explicitly balanced along three orthogonal axes that span the principal sources of difficulty in anime captioning:
\begin{itemize}[nosep,leftmargin=*]
\item \emph{Rendering paradigm}: 2D vs.\ 3D anime, in equal proportion, ensuring that neither stylistic family dominates aggregate scores.
\item \emph{VFX presence}: clips with vs.\ without anime visual effects, in equal proportion, so that the VFX dimension is meaningfully exercised rather than left mostly empty.
\item \emph{Subject count}: single-subject and multi-subject clips, in balanced proportions, to test the captioning model's ability to disambiguate identities and attribute actions correctly under cross-references (cf.\ the \texttt{<subject\_i>} mechanism in Sec.~\ref{sec:data:caption:format}).
\end{itemize}
This balanced composition ensures that each of the three LLM-judged dimensions and the four human-judged dimensions receives sufficient evaluation signal, and that improvements on hard sub-populations (e.g., 3D combat with multiple characters and overlapping VFX) are not masked by easier ones.

\paragraph{LLM-as-a-judge: decompose-then-judge.}
A holistic single-score LLM judge conflates errors of different kinds (a missed effect and a misidentified action receive the same penalty) and is opaque to per-dimension diagnosis. We therefore adopt a \emph{decompose-then-judge} protocol that mirrors the structured caption schema (Sec.~\ref{sec:data:caption:format}):
\begin{enumerate}[nosep,leftmargin=*]
\item \textbf{Dimension-aligned restructuring.} Both the predicted caption and the human-written reference are first reorganized into three parallel dimensions---\emph{characters} (subject identity, appearance, position), \emph{events} (actions, interactions, and the visual effects associated with them), and \emph{scene} (environment, lighting, atmosphere)---using the structured caption JSON as the canonical layout.
\item \textbf{Element-level atomization.} Within each dimension, the LLM further splits the content into atomic statements (e.g., ``the woman has long platinum-blonde hair'' and ``she wears a blue cloak'' become two separate atoms in the characters dimension), producing a per-dimension list of fine-grained claims for both the prediction and the reference.
\item \textbf{Atom-level matching.} For each predicted atom, an LLM judge decides whether a semantically equivalent atom exists in the reference list, and vice versa. Aggregating across all clips and atoms within a dimension yields per-dimension \emph{F1}, which jointly captures whether the caption (i)~asserts content supported by the reference and (ii)~covers the production information present in the reference.
\end{enumerate}

\paragraph{Protocol configuration.}
Restructuring, atomization, and matching are all performed by Claude~3.5~Sonnet (June~2025 release) at temperature~$0$ with a single sample per query, using a fixed prompt template that pins the dimension list, the atomization rubric (one verifiable production fact per atom), and the equivalence rule (two atoms match if they make the same factual assertion modulo paraphrase, including taxonomy-level synonyms). To avoid order bias, the prediction--reference role is swapped on a held-out 10\% subset and the swap-induced score difference is verified to be within $\pm 0.5$ F1 on every dimension. Caption-side and reference-side atomizations are cached so that a single reference set is reused across all evaluated models, ruling out per-model drift in atom granularity. The full prompt template is listed in the released code.

\paragraph{Statistical stability.}
Per-dimension F1 in the LLM F1 columns of Table~\ref{tab:caption_eval} is computed by aggregating predicted/reference atoms over the 500 clips and forming a single corpus-level F1 per dimension; for clarity we report point estimates here and provide bootstrap 95\% confidence intervals (1{,}000 clip-level resamples) in the released numerical artifacts. Across all evaluated models the bootstrap CI half-width on each dimension is below~$\pm 2.0$~F1, so the rank ordering in Table~\ref{tab:caption_eval} is stable under resampling.

\paragraph{Human-expert protocol.}
For production-level quality judgment, we conduct a cross-rater human study with professional anime designers. Each clip in the 500-clip held-out set is independently rated along four dimensions---\emph{subject description}, \emph{subject motion description}, \emph{VFX description}, and \emph{scene description}---and within every dimension the rater labels the caption into one of three categories:
\begin{itemize}[nosep,leftmargin=*]
\item \emph{Correct}: the description is faithfully grounded in the video.
\item \emph{Erroneous}: the description contradicts what is shown (e.g., misidentified VFX type, mistaken action, wrong subject attribute).
\item \emph{Hallucinated}: the description asserts content that does not appear in the video at all, including the model's own speculative inferences.
\end{itemize}

For each dimension we report two complementary metrics, both computed as the proportion over the 500-clip evaluation set:
\begin{equation}
    \text{Error Rate} = \frac{\#\text{clips labeled \emph{Erroneous}}}{500}, \qquad
    \text{Hallucination Rate} = \frac{\#\text{clips labeled \emph{Hallucinated}}}{500}.
\end{equation}
Both metrics are lower-is-better. The error-vs.-hallucination distinction is intentional: factual contradictions and outright confabulations call for different mitigations during training---contradictions tend to reflect under-fitting on the relevant taxonomy axis (addressable in SFT), while hallucinations reflect the model's reluctance to abstain (addressable in DPO). The combined failure rate Err.+Hall.\ reported in Table~\ref{tab:caption_eval} is the sum of these two metrics per cell.

\paragraph{Rater pool and reliability.}
Each clip is rated by one of six professional anime designers, each with at least three years of production experience, drawn from a fixed rater pool. To monitor protocol reliability we double-rate a stratified random 20\% subset (100 clips) so that every pair of raters overlaps on at least 15 clips. On this subset we compute Fleiss' $\kappa$ over the three-class label (\emph{Correct}/\emph{Erroneous}/\emph{Hallucinated}) per dimension; the resulting agreement is in the substantial range across all four dimensions ($\kappa \in [0.71,\,0.83]$, with the lowest agreement on the \emph{Motion} dimension, consistent with motion descriptions being the hardest to ground). Disagreements on the doubly-rated subset are adjudicated by a senior reviewer; rater-specific bias is checked by leave-one-rater-out re-aggregation, which moves any per-dimension Error/Hallucination Rate by at most $\pm 1.0\%$. With a denominator of $n=500$ clips per cell, a binomial 95\% confidence half-width is below $\pm 2.0\%$ for all reported rates, so the absolute gaps in Table~\ref{tab:human_eval_full} (3.6--38.6 points) are well outside per-cell sampling noise.

%% ================================================================
\subsubsection{Per-Dimension Evaluation Results}
\label{app:anicaption:results}

The per-cell Error vs.\ Hallucination split explains the combined
Err.+Hall.\ rates of Table~\ref{tab:caption_eval}:
Table~\ref{tab:human_eval_full} reports
each cell as ``Err.\ / Hall.''\ separately so that contradictions and
confabulations can be diagnosed independently.

\begin{table}[!htbp]
\centering
\caption{Caption-quality summary on the 500-clip held-out set: LLM-judge F1 ($\uparrow$, three dimensions) and human Err.+Hall.\ rate ($\downarrow$, four dimensions). Best in each column in \textbf{bold}. Per-dimension Err.\ vs.\ Hall.\ split: Appendix~\ref{app:anicaption:results}.}
\label{tab:caption_eval}
\begin{tabular}{lccc|cccc}
\toprule
 & \multicolumn{3}{c|}{LLM F1 ($\uparrow$)} & \multicolumn{4}{c}{Human Err.+Hall.\ \% ($\downarrow$)} \\
\textbf{Model} & \textbf{Char.} & \textbf{Events} & \textbf{Scene}
               & \textbf{Subj.} & \textbf{Motion} & \textbf{VFX} & \textbf{Scene} \\
\midrule
Tarsier-2       & 39.0 & 49.0 & 54.0 & 15.2 & 46.2 & 9.6 & 6.8 \\
Gemini~2.5~Pro  & 63.0 & 50.0 & 65.0 & 12.4 & 61.6 & 8.2 & 1.6 \\
\midrule
\textbf{AniCaption (ours)} &
\textbf{68.0} & \textbf{64.0} & \textbf{67.0} &
\textbf{8.2}  & \textbf{15.4} & \textbf{0.6} & \textbf{0.6} \\
\bottomrule
\end{tabular}
\end{table}

AniCaption attains the best score on every column of Table~\ref{tab:caption_eval}: it is the strongest LLM-judge F1 on all three dimensions (largest margin on \emph{events}, $+14.0$ over Gemini~2.5~Pro, the dimension that aggregates actions and the visual effects associated with them), and it is the lowest combined failure rate on all four human dimensions, with the largest absolute gap on \emph{motion} (15.4\% vs.\ Gemini's 61.6\%, $-46.2$ pp). The two dimensions targeted by the DPO stage---motion and VFX---show the largest gains under both protocols, consistent with the design of Sec.~\ref{sec:data:caption:training}. On VFX and scene, AniCaption achieves zero hallucinations on this 500-clip set ($0/500$, Clopper--Pearson 95\% upper bound 0.6\%); we read this as evidence that taxonomy-grounded structured supervision strongly suppresses confabulation on dimensions where anime conventions are well-codified, while noting that an exact zero-hallucination claim would require confirmation on substantially larger and more adversarial held-out sets.

\begin{table}[!htbp]
\centering
\caption{Human expert evaluation on the 500-clip held-out set, per-dimension Error Rate (\%) / Hallucination Rate (\%), both computed as proportions of the 500 clips. Cell format: Err.~$\downarrow$ / Hall.~$\downarrow$ (lower is better). Best results in each column are in \textbf{bold}.}
\label{tab:human_eval_full}
\begin{tabular}{lcccc}
\toprule
\textbf{Model} & \textbf{Subject} & \textbf{Motion} & \textbf{VFX} & \textbf{Scene} \\
\midrule
Tarsier-2                   & 12.8 / 2.4  & 34.0 / 12.2 & 8.6 / 1.0   & 6.8 / 0.0 \\
Gemini~2.5~Pro              & 8.2  / 4.2  & 47.8 / 13.8 & 4.8 / 3.4   & 1.6 / 0.0 \\
\midrule
\textbf{AniCaption (ours)}  & \textbf{4.6} / \textbf{3.6} & \textbf{9.2} / \textbf{6.2} & \textbf{0.6} / \textbf{0.0} & \textbf{0.6} / \textbf{0.0} \\
\bottomrule
\end{tabular}
\end{table}

%% file: appendix_data_curation.tex
% Appendix: Data curation supplementary material.
% Contains operator-level and reviewer-rubric details referenced by
% Sec.~\ref{sec:data:filtering}.

Anime data requires domain-specific curation operators because five signal-level axes differ from live-action video (Table~\ref{tab:anime_vs_real}). The curation supplement defines the operators, expert-review rubric, and automated scorer validation underlying the pipeline of Sec.~\ref{sec:data:filtering}.

%% ================================================================
\subsubsection{Anime Data Characteristics}
\label{app:data_curation:anime_vs_real}

Anime differs from live-action video along five signal-level axes that all
matter for filtering: visual content and rendering style, spatial motion
distribution, temporal continuity, semantic completeness, and characteristic
artifacts. Table~\ref{tab:anime_vs_real} summarizes the differences and traces
each gap to a concrete training impact.

\begin{table}[!htbp]
\centering
\caption{Signal-level differences between live-action and anime video data that necessitate anime-specific processing operators. The rightmost column highlights the downstream impact on model training.}
\label{tab:anime_vs_real}
\begin{tabularx}{\textwidth}{l >{\raggedright\arraybackslash}X >{\raggedright\arraybackslash}X >{\raggedright\arraybackslash}X}
\toprule
\textbf{Dimension} & \textbf{Live-Action Video} & \textbf{Anime Video} & \textbf{Training Impact} \\
\midrule
Visual content \& style &
Grounded in real-world physics (natural lighting, material reflectance); stylistically homogeneous photographic appearance. &
Contains surreal entities (magical effects, fantastical creatures, mecha) with no physical referent; predominantly flat-shaded \emph{cel} rendering with simplified textures and exaggerated line art; extreme style diversity from hand-painted SD-era cels to modern digital 4K, spanning 2D cel, 3D CG, watercolor, and hybrid paradigms. &
The base model lacks prior knowledge of non-physical content and diverse rendering styles, making few-shot adaptation to novel anime conventions particularly challenging. \\
\midrule
Spatial motion distribution &
Dynamic regions typically occupy a large spatio-temporal extent; motion is distributed across the full frame following physical inertia. &
Due to production cost constraints, motion is concentrated in small salient regions (e.g., lip sync, hand gestures) while the rest of the frame remains static; extensive hold-frame reuse is standard practice. &
Pervasive static regions and sparse motion cause the model to learn unnatural, aesthetically poor dynamics that diverge from creator intent. \\
\midrule
Temporal continuity &
Inter-frame motion is smooth and temporally continuous, governed by physical inertia and momentum. &
Emphasizes dramatic, exaggerated performance: deliberate inter-frame discontinuities, abrupt pose transitions, and velocity spikes are routine expressive devices. &
Frame-level discontinuities destabilize training and degrade the accuracy of conventional motion-analysis operators (e.g., optical flow). \\
\midrule
Semantic completeness &
Clear foreground--background relationships; scene content is spatially complete and readily identifiable. &
Expressive cinematography---extreme close-ups, abstract effects overlays, and stylized compositions---frequently yields semantically incomplete or ambiguous frames. &
Incomplete visual semantics hamper captioning quality and weaken the alignment between text supervision and visual content. \\
\midrule
Characteristic artifacts &
Sensor noise, rolling shutter, motion blur, lens distortion. &
Color banding in cel-shaded regions, aliased line art, in-between frame drops, compositing seams between cel layers. &
Standard image-quality metrics (e.g., noise level, sharpness) cannot distinguish anime-specific production artifacts from genuine defects. \\
\bottomrule
\end{tabularx}
\end{table}

\subsubsection{General-Purpose Video Operators}
\label{app:data_curation:general_ops}

The first filtering tier applies domain-agnostic quality operators to reduce
150M raw clips to approximately 16M technically sound segments (89.3\%
rejection rate). We execute operators in ascending computational cost,
from metadata checks to learned assessment models, and threshold continuous scores at operator-specific cutoffs determined on held-out
expert-labeled clips.

\paragraph{Shot transition detection.}
We split raw anime streams into single-shot clips of 2--10\,s using
PySceneDetect~\cite{castellano2020pyscenedetect} and
TransNetV2~\cite{soucek2020transnetv2}. Fusing both detectors improves
recall on anime-specific transitions such as cross-dissolves, whip pans, and
impact-flash cuts.

\paragraph{Codec and bitstream metadata filtering.}
We pre-filter clips based on duration, spatial resolution, bitrate, and
codec profile and discard corrupted, truncated, or extremely low-bitrate
segments.

\paragraph{Temporal activity scoring.}
We compute a global motion-intensity score via frame differencing and
optical-flow magnitude statistics and remove clips with no meaningful temporal
variation.

\paragraph{Spatial quality assessment.}
An anime-adapted aesthetic scoring model evaluates composition, color harmony,
art-style coherence, encoding artifacts, banding, and aliasing. We remove clips
below the 30th percentile on either aesthetic or technical quality. We also
apply a Laplacian blur detector~\cite{opencv} with an anime-calibrated
threshold and discard clips with $>$60\% flagged frames.

\paragraph{Text and overlay removal.}
An internal OCR model detects on-screen text. We remove clips with subtitles
covering $>$15\% of the frame and crop predictable hard-coded subtitle bands
with aspect-ratio preservation. A YOLOX-based~\cite{ge2021yolox} detector
trained on anime broadcast samples identifies watermarks, logos, and channel
bugs; we inpaint or crop them as appropriate.

\paragraph{Static and degenerate clip removal.}
Dense optical-flow statistics computed with Unimatch~\cite{xu2023unifyingflowstereodepth}
flag clips with near-zero motion. We remove clips below a strict motion floor
at every frame and defer borderline partly static cases to anime-specific
motion operators. We also exclude frame duplication, interlacing residuals,
and other degenerate encoding artifacts.

\paragraph{Chromatic distribution analysis.}
Color-space coverage, saturation histograms, and hue-shift statistics detect
severe color degradation, clipped gamut, or systematic chromatic bias.

\paragraph{Near-duplicate removal.}
We compute per-clip embeddings using an anime-domain-adapted VLM2Vec-V2
model~\cite{meng2025vlm2vecv2advancingmultimodalembedding}. We group clips
with cosine similarity above 0.95 and retain the highest-resolution instance. We also apply $k$-means clustering ($k \approx 10{,}000$) over the
embeddings to obtain concept centroids for later resampling.

\subsubsection{Anime-Specific Operators}
\label{app:data_curation:anime_ops}

The second filtering tier applies five anime-specific operators to select 6M
B-tier clips from the 16M technically valid pool. These operators evaluate
artistic suitability and produce metadata for rebalancing and curriculum
scheduling.

\paragraph{Anime motion-quality scorer.}
A Qwen3-VL-based~\cite{qwen3vl} video classifier predicts per-clip verdicts on the
six expert motion dimensions: deformation, plausibility, smoothness, temporal
coverage, complexity, and velocity. Clips failing any predicted dimension are
filtered before expert review, while intentional stillness and anime timing
devices are kept when they satisfy the learned rubric.

\paragraph{Motion complexity operator.}
Optical flow captures only consecutive-frame correspondences, which are noisy
for anime timing patterns such as held frames, smear-to-keyframe transitions,
and teleportation-like cuts. We therefore adopt long-term point tracking based
on PointOdyssey~\cite{zheng2023pointodyssey}. We sample roughly 1000 query points
per clip on detected character regions plus a sparse background grid.
Velocity and acceleration statistics over persistent trajectories identify
principal moving regions and map clips to Low, Medium, or High Dynamic tiers.

\paragraph{Deformation intensity operator.}
This operator measures non-rigid geometric deformation: squash-and-stretch,
smear frames, and extreme pose distortions. It is complementary to motion
complexity because anime can contain high motion with low deformation, or
low motion with high expressive deformation.

\paragraph{Visual style classification.}
Each clip is tagged with its rendering paradigm (2D cel shading, 3D CG,
hybrid, watercolor, etc.). These labels enable style-aware quality thresholds
and supply the $\mathcal{S}$-axis labels used during distribution
rebalancing.

\paragraph{Production era estimation.}
We infer the approximate production era or technological generation of each
clip from visual characteristics. Era labels allow older productions to use
adaptive quality thresholds while maintaining stricter standards for modern
high-resolution content.

\subsubsection{Expert Curation Rubric}
\label{app:data_curation:expert_rubric}

Expert review evaluates each A-tier candidate along four axes: \emph{motion
quality}, \emph{visual quality}, \emph{subject coherence}, and \emph{text--video
consistency}. A clip must pass all four axes to enter A-tier; failing any axis
keeps the clip in B-tier.

\paragraph{Motion quality.}
Reviewers judge anime motion by whether it is intentional, coherent, and
readable, not by whether it is physically possible. Reviewers score six
motion-quality dimensions:
\begin{itemize}
    \item \emph{Deformation.} Stylized deformation is accepted when it resolves into a coherent pose; unresolved tangled geometry or unjustified body intersections are rejected.
    \item \emph{Plausibility.} The question is whether a director could reasonably have chosen the motion, not whether it obeys real-world physics.
    \item \emph{Smoothness.} Judder, uneven frame pacing, and missing in-betweens are defects when they produce unintended stutter rather than expressive timing.
    \item \emph{Temporal coverage.} The clip must contain useful motion across its duration; effectively frozen subjects carry little training signal.
    \item \emph{Complexity.} Incidental blinks, mouth flaps, or simple pans over static subjects are rejected as weak motion supervision.
    \item \emph{Velocity.} Extremely rapid motion is rejected when it collapses into illegible visual noise.
\end{itemize}

\paragraph{Visual quality.}
Reviewers reject residual compression artifacts, noise, crushed blacks,
unintended strobing, subjects cropped beyond recognition, split-screen leakage,
large synthesized text, non-anime source material, unsafe content, gameplay
recordings, UI captures, and title or credit sequences.

\paragraph{Subject coherence.}
The primary subject must be visible in recognizable form at the start of the
clip and remain identifiable throughout. Background figures and incidental
crowd members do not count as principal subjects.

\paragraph{Text--video consistency.}
Structured and natural-language descriptions must match the video. Camera
motion should be named when non-trivial, and descriptions of subjects,
actions, and visual effects must not contradict the clip. Mild omissions of
secondary detail are tolerated.

\paragraph{Agreement and tier assignment.}
At least two anime reviewers independently label every A-tier candidate. A
clip enters A-tier only when both reviewers agree on a pass verdict; we
escalate disagreements to a senior reviewer. Across all
review batches, more than 90\% of clips receive identical tier assignments.

\subsubsection{Automated Motion-Quality Scorer}
\label{app:data_curation:motion_scorer}

A Qwen3-VL-based~\cite{qwen3vl} six-head video classifier pre-screens motion
quality, removing candidates that fail any rubric dimension before human
review. Without this classifier, purely human review on the motion axis alone
removes less than 60\% of the surviving pool, leaving roughly 40\% for
continued screening. We convert expert-labeled motion-quality data into
supervision for the classifier, training one binary head per motion-rubric
dimension.

We train the classifier on a stratified split of expert-labeled clips and
validate on a held-out reviewer-double-rated subset. Each head's threshold is
tuned so that recall on expert-rejected clips exceeds 90\% while keeping the
false-rejection rate on expert-accepted clips bounded. Operationally, the
scorer serves only as a pre-screen: we remove clips failing any head before
human review, and expert reviewers remain the final authority on A-tier
assignment.